\documentclass[runningheads]{llncs}
\usepackage{graphicx}
\usepackage{subcaption} % Add By Qiankun Gao
\usepackage{tikz}
\usepackage{comment}
\usepackage{amsmath,amssymb} % define this before the line numbering.
\usepackage{color}
\usepackage[breaklinks=true,
            bookmarks=false,
            colorlinks,
            linkcolor=red,       
            anchorcolor=blue,  
            citecolor=green, 
            ]{hyperref}
\usepackage[accsupp]{axessibility}  % Improves PDF readability for those with disabilities.

\usepackage{dsfont} % Add By Qiankun Gao
\usepackage{booktabs} % ADD BY Qiankun Gao
\usepackage{multirow} % ADD BY Qiankun Gao
\usepackage{pifont} % ADD BY Qiankun Gao
\usepackage{xspace}
\makeatletter
\DeclareRobustCommand\onedot{\futurelet\@let@token\@onedot}
\def\@onedot{\ifx\@let@token.\else.\null\fi\xspace}

\def\eg{\emph{e.g}\onedot} 
\def\ie{\emph{i.e}\onedot} 
 
 \def\vs{\emph{vs}\onedot}
 
\def\etal{\emph{et al}\onedot}
\makeatother
\begin{document}
% \linenumbers
\pagestyle{headings}
\mainmatter
\def\ECCVSubNumber{3118}  % Insert your submission number here

\newcommand{\chen}[1]{{{\color{red}[\textbf{Chen}: \emph{#1}]}}}

\title{R-DFCIL: Relation-Guided Representation Learning for Data-Free Class Incremental Learning} % Replace with your title

\begin{comment}
\titlerunning{ECCV-22 submission ID \ECCVSubNumber} 
\authorrunning{ECCV-22 submission ID \ECCVSubNumber} 
\author{Anonymous ECCV submission}
\institute{Paper ID \ECCVSubNumber}
\end{comment}
%******************

%\begin{comment}
\titlerunning{R-DFCIL: Relation-Guided Representation Learning for DFCIL}
\author{
Qiankun Gao\inst{1}%\orcidID{0000-1111-2222-3333} 
\and
Chen Zhao\inst{2}%\orcidID{1111-2222-3333-4444}
\and
Bernard Ghanem\inst{2}%\orcidID{1111-2222-3333-4444}
\and
Jian Zhang\inst{1}\thanks{Corresponding author: Jian Zhang.}%\orcidID{1111-2222-3333-4444}
}
\authorrunning{Q. Gao et al.}
\institute{
Peking University Shenzhen Graduate School, Shenzhen, China \\
\email{gqk@stu.pku.edu.cn, zhangjian.sz@pku.edu.cn}
%\url{https://www.ece.pku.edu.cn}
\and
King Abdullah University of Science and Technology (KAUST), Thuwal, Saudi Arabia \\
\email{\{chen.zhao,bernard.ghanem\}@kaust.edu.sa}
}
%\end{comment}
%******************
\maketitle

\begin{abstract}
Class-Incremental Learning (CIL) struggles with catastrophic forgetting when learning new knowledge, and Data-Free CIL (DFCIL) is even more challenging without access to the training data of previously learned classes. Though recent DFCIL works introduce techniques such as model inversion to synthesize data for previous classes, they fail to overcome forgetting due to the severe domain gap between the synthetic and real data. To address this issue, this paper proposes relation-guided representation learning (RRL) for DFCIL, dubbed R-DFCIL. In RRL, we introduce relational knowledge distillation to flexibly transfer the structural relation of new data from the old model to the current model. Our RRL-boosted DFCIL can guide the current model to learn representations of new classes better compatible with representations of previous classes, which greatly reduces forgetting while improving plasticity. To avoid the mutual interference between representation and classifier learning, we employ local rather than global classification loss during RRL. After RRL, the classification head is refined with global class-balanced classification loss to address the data imbalance issue as well as learn the decision boundaries between new and previous classes. Extensive experiments on CIFAR100, Tiny-ImageNet200, and ImageNet100 demonstrate that our R-DFCIL significantly surpasses previous approaches and achieves a new state-of-the-art performance for DFCIL. Code is available at \href{https://github.com/jianzhangcs/R-DFCIL}{\texttt{https://github.com/jianzhangcs/R-DFCIL}}

\end{abstract}

\section{Introduction}
Class-Incremental Learning (CIL) is a learning paradigm in which a model (referred to as a solver model) continually learns a sequence of classification tasks. 
The model suffers from catastrophic forgetting~\cite{catastrophic_forgetting,cfrp} since its access to data of previous tasks is restricted when learning a new task.
Existing CIL works~\cite{icarl,ucir,podnet,e2eil,aanets} try to overcome the challenge mainly through saving a small proportion of previous training data in memory. 
Despite their success of mitigating catastrophic forgetting, these approaches may bring issues such as violation of data legality and  explosion of storage space.
Instead, some works~\cite{dgr,ganmemory,fearnet} simultaneously train the solver model and a data generator, which is used to generate data for previous classes at a new task. 
This usually performs poorly and still causes data privacy concerns because the generator may remember sensitive information in the real data.
To address these concerns, researchers start to consider Data-Free CIL (DFCIL)~\cite{lwf,deepinversion,abd}, 
in which the model incrementally incorporates new information without storing data or generator of previous tasks.

Early DFCIL works, \eg, LwF~\cite{lwf}, are often ineffective in overcoming catastrophic forgetting without data of previous tasks~\cite{bircl}. More recently, Yin~\etal~introduce model inversion~\cite{deepinversion} 
to DFCIL to synthesize data for previous tasks when learning a new task, the forgetting of previous classes can be mitigated by performing knowledge distillation on these synthetic data.
However, the synthetic data have a severe domain gap with the real data, misleading the decision boundaries between new and previous classes.
These approaches may come through the first few tasks (\ie, short-term CIL), but they lose the stability-plasticity balance 
when learning many tasks (\ie, long-term CIL). It is still a great challenge to train a model with both good stability (\ie, not forgetting previous knowledge) and plasticity (\ie, learning new knowledge) in DFCIL.

After a thorough study on DFCIL with synthetic data of previous classes, we identify bottlenecks in prior approaches as follows: 
\textbf{1)} with the existence of domain gap between synthetic and real data, the global classification loss (\ie, the cross-entropy between the model's prediction among all seen classes and the ground truth) leads classifiers to separate new and previous classes by domain features rather than semantic features, which also causes the model to learn more domain features of synthetic data than semantic features of previous classes;
\textbf{2)} to overcome forgetting, prior works perform the same knowledge distillation method on the synthetic data and the data of new classes, ignoring the difference between them, which actually hurts the model's plasticity and is not helpful in alleviating the conflict between improving plasticity and maintaining stability. Please refer to the supplementary material for more details. 

To address the above bottlenecks, we propose 
\textbf{1)} relation-guided representation learning (RRL) with hard knowledge distillation (HKD) for synthetic old data together with the relational knowledge distillation (RKD) for data of new task; 
\textbf{2)} local classification loss (\ie, the cross-entropy between the model's prediction among new classes and ground truth) in place of global classification loss during representation learning, following classification head refinement with global class-balanced classification loss using a small learning rate.

Specifically, our novel approach R-DFCIL consists of three stages: 
\textbf{1)} before learning a new task, we \textbf{train an image synthesizer} by inverting the old model through model inversion technique~\cite{deepinversion}, which is used to synthesize image during learning new task; 
\textbf{2)} we design three components to encourage the model to learn the representations of new classes without forgetting learned classes, in which \textbf{local classification loss} improves model's plasticity, \textbf{hard knowledge distillation} maintains model's stability, and \textbf{relational knowledge distillation} mitigates the conflict between them; 
\textbf{3)} after representation learning, we refine the classification head to address the data imbalance between classes as well as learn the decision boundaries between new and previous classes, in which a \textbf{global class-balanced classification loss} is adopted, and the weights of classes are computed by their number of training samples.

We summarize our contributions as follows:
\begin{itemize}
\item We propose a novel DFCIL approach R-DFCIL, which strikes a better stability-plasticity balance by relation-guided representation learning (RRL) and classification head refinement (CHR).
\item To the best of our knowledge, we are the first to introduce relational knowledge distillation (RKD) to DFCIL, which is critical to mitigate the conflict between learning the representations for new classes and preserving the representations of previously learned classes.
\item We conduct extensive experiments on CIFAR100~\cite{cifar} , Tiny-ImageNet200~\cite{tiny_imagenet}, and ImageNet100~\cite{ucir} datasets, on all of which, our R-DFCIL surpasses the previous state-of-the-art ABD~\cite{abd}
with accuracy gains of 8.46\%, 9.23\%, and 9.88\%, respectively,
and achieves a new record for DFCIL. 
\end{itemize}

\section{Related Work}
\label{sec:related_work}

\smallskip\noindent\textbf{Class-Incremental Learning (CIL).} To overcome catastrophic forgetting, successful approaches~\cite{icarl,e2eil,ucir,sdc,podnet,gdumb,aanets,rainbow} store representative training data for previously learned classes and replay them when updating the model with the data from new task. 
Knowledge distillation (KD)~\cite{knowledge_distillation} techniques are widely used in these approaches to further alleviate forgetting of learned information, \eg, iCaRL~\cite{icarl} conducts KD on the pre-softmax output of the old and new data, UCIR~\cite{ucir} designs a novel feature distillation loss, and PODNet~\cite{podnet} proposes to distill from not only the final embedding output but also the pooled output of the model's intermediate layers. 
However, these methods are not suitable for synthetic data, so we adopt a hard KD, which directly distills the knowledge from the model's output. 
PODNet requires another stage to fine-tune the classifier with balanced data, our approach also has a classification head refinement stage, in which the model addresses the data imbalance issue and learns decision boundaries between new and previous classes with a global class-balanced classification loss. 
The classification head also impacts the incremental performance: iCaRL works better with NME than CNN classifier, UCIR is more compatible with cosine classifier, and PODNet depends on LSC classifier. We remove the bias parameter of the linear classifier to adapt our approach better.

\smallskip\noindent\textbf{Data-Free Class-Incremental Learning (DFCIL).} The earliest DFCIL work is LwF~\cite{lwf}, which first introduces knowledge distillation (KD) to incremental learning. Unfortunately, KD has limited effectiveness in overcoming forgetting when using only new data. 
Some prior works~\cite{dgr,ganmemory,fearnet,lvaegan,mrgan} train a large generator simultaneously with the training of the solver model, which helps the solver model remember the knowledge of previous tasks through replaying the generated data.
These approaches usually perform poorly~\cite{bircl} due to the domain gap between generated and real data, and they also cause data privacy concerns because the generator may remember sensitive information in the real data~\cite{mmingan}. 
Recent works~\cite{deepinversion,abd} introduce model inversion technique to synthesize data of previous tasks. Although the visual quality is very different from the real images, the synthetic images generally match the statistical distribution of the real data from previous tasks. The synthetic images often mix features from multiple classes, which confuse the decision boundaries between classes, the prior approaches that overcome forgetting with real data may fail with synthetic data. Our approach follows ABD~\cite{abd} to synthesize data for previous classes by model inversion technique, but we further propose a training framework that separates representation and classifier learning to avoid the mutual interference caused by domain gap between synthetic and real data.

\smallskip\noindent\textbf{Knowledge Distillation (KD)} was first introduced to Deep Learning by Hinton \etal~\cite{knowledge_distillation} to transfer knowledge from a teacher model to a small student model. Since then, various KD methods~\cite{fitnets,svdkd,jmkd,pkt} have been developed. Conventional KD methods extract knowledge from individual data, \ie, keep the hidden activation or the final output of the student model consistent with those of the teacher model for individual training samples. In contrast, Park \etal.~\cite{rkd} propose Relational KD (RKD) to transfer structural knowledge using mutual relations of data examples in the teacher’s output presentation. Their experimental results demonstrate that RKD is superior to conventional individual KD (IKD) methods. KD techniques are also widely used in incremental learning to overcome catastrophic forgetting, but most of them are IKD methods. These IKD methods can improve the model's stability when applied to old data but may hurt the model's plasticity when applied to new data. Inspired by RKD, we propose relation-guided representation learning to address DFCIL problem. 
\section{Methodology}
\label{sec:methodology}

\subsection{Problem formulation and R-DFCIL architecture}
\subsubsection{Problem formulation.} 
In the problem of Data-Free Class-Incremental Learning (DFCIL), a model sequentially learns a series of tasks, in which the $i^{th}$ task introduces a set of classes $\mathcal{T}_i$ that do not overlap with those in previous tasks. 
We use $\mathcal{T}_i$ and the $i^{th}$ task interchangeably in this paper, and denote the number of classes in $\mathcal{T}_i$ as $|\mathcal{T}_i|$.
At learning phase $i$, the model can only access the training data of the current task $\mathcal{T}_i$, and  predicts for all the data of the tasks $\mathcal{T}_{1:i}$ (\ie, from $\mathcal{T}_1$ to $\mathcal{T}_i$) for inference after the learning is finished. 
We denote the feature extractor with stacks of convolutional layers as $f: \mathbb{R}^{h \times w \times 3} \to \mathbb{R}^d$, and the classification head with $c$ linear classifiers as $\theta: \mathbb{R}^d \to \mathbb{R}^c$, then the model  $\theta \circ f$  predicts the  class $y$ of  input $\mathbf{x}$ via $\hat{y} = \mathop{\arg \max}_{j \in \{0, \dots, c - 1\}} \theta^{(j)}(f(\mathbf{x}))$. 
For simplicity, we denote the frozen snapshot of $\theta \circ f$ at the end of learning phase $i$ as $\theta_{i} \circ f_{i}$, which means $\theta_{i} \circ f_{i}$ has learned $\mathcal{T}_{1:i}$. 
The training data and test data of $\mathcal{T}_i$ are described by $\mathcal{D}_i^{train}$ and $\mathcal{D}_i^{test}$, respectively. We also refer $\mathcal{D}_{1:i}^{train}$ and $\mathcal{D}_{1:i}^{test}$ to the training and test data of $\mathcal{T}_{1:i}$ for convenience. 

\begin{figure}[t]
\centering
\includegraphics{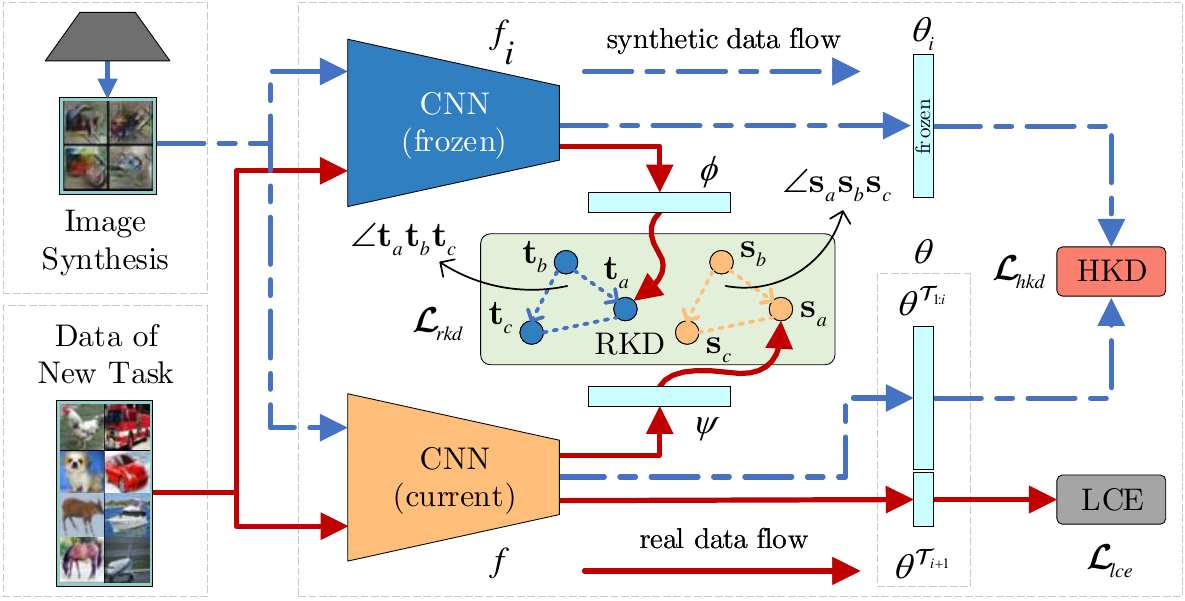}
\caption{Overview of our R-DFCIL. The model $\theta \circ f$ is learning the current task $\mathcal{T}_{i+1}$.
The hard knowledge distillation loss $\mathcal{L}_{hkd}$
is applied on synthetic data to alleviate forgetting. The local classification loss $\mathcal{L}_{lce}$ is employed on new data to learn new knowledge. The relation knowledge distillation loss $\mathcal{L}_{rkd}$ transfers the structural relation of new data $\mathcal{D}_{i+1}^{train}$ from the previous model $\theta_i \circ f_i$ to the current model $\theta \circ f$. 
$\phi$ and $\psi$ are two linear transform functions. 
\vspace{-10pt}
} 
\label{fig:rgrl}
\end{figure}

\subsubsection{R-DFCIL architecture.} 
Fig.~\ref{fig:rgrl} illustrates the architecture of our relation-guided representation learning for DFCIL (R-DFCIL). Our R-DFCIL is based on the framework that synthesizes old data when learning a new task, and contains the following  three stages:
\textbf{First}, at the beginning of the learning phase $i$$+$$1$, we  train a synthesizer by inverting the old model $\theta_{i} \circ f_{i}$ through model inversion technique~\cite{deepinversion} following ABD~\cite{abd}. 
\textbf{Then}, the model starts to learn new task $\mathcal{T}_{i+1}$ once the synthesizer training is completed. We temporarily keep the snapshot $\theta_{i} \circ f_{i}$ (old model) in memory, and add $|\mathcal{T}_{i+1}|$ new linear classifiers (denoted as $\theta^{\mathcal{T}_{i+1}}$) to $\theta$ (the original $\theta$ is denoted as $\theta^{\mathcal{T}_{1:i}}$). Then, we randomly sample a batch of training data $(X^{new}, Y^{new})$ from the new training data $\mathcal{D}_{i+1}^{train}$, and synthesize the same number of data $(X^{old}, Y^{old})$ by the synthesizer for previous classes, which are passed to the model to learn the representations of new classes without forgetting previous classes by integrating the hard knowledge distillation, local classification loss and relational knowledge distillation. 
\textbf{Last}, after representation learning, we freeze the feature extractor $f$ and refine the classification head $\theta$ with global class-balanced classification loss to address the data imbalance issue as well as learn the decision boundaries between new and previous classes. 

In the following subsections, we will first describe our core contributions of relation-guided representation learning in Sec.~\ref{sec:RRL} and classification head refinement in Sec.~\ref{sec:CHR}, then review the synthesizer training in Sec.~\ref{sec:IS}.

\subsection{Relation-Guided Representation Learning}\label{sec:RRL}
Learning new knowledge will inevitably change the current model, causing the forgetting of previously learned classes. Therefore, on the one hand, how to overcome forgetting is essential in DFCIL. To this end, we provide the technique of hard knowledge distillation (HKD). On the other hand, the model should also have the flexibility to learn knowledge from the classes in the new task, for which we adopt local cross-entropy loss (LCE) on data of new classes. However, the conflict between overcoming forgetting by HKD and learning new knowledge by LCE still can not be well resolved, which motivates us to propose relation-guided representation learning (RRL) via relational knowledge distillation (RKD).  

\smallskip\noindent\textbf{Hard Knowledge Distillation (HKD).} 
Prior works usually take the following knowledge distillation method to keep the model from forgetting previous $1:i$ tasks when learning the $i$$+$$1^{\textrm{th}}$ task:
\begin{equation}
\mathcal{L}_{kd} = \frac{1}{|X|} \sum_{\mathbf{x} \in X} \mathcal{D}_{KL}\left(
\operatorname{softmax}\left(
\theta_i \left( f_{i} \left( \mathbf{x} \right ) \right) / \tau
\right),
\operatorname{softmax}\left(
\theta^{\mathcal{T}_{1:i}} \left( f \left( \mathbf{x} \right ) \right) / \tau
\right)
\right),
\end{equation}
where $\tau$ is a temperature parameter, $\mathcal{D}_{KL}$ is KL divergence, and $X$ is one of $X^{new}$, $X^{old}$ and $X^{new} \cup X^{old}$. However, we find that it is not hard enough when applied to synthetic data.
Instead, we use a harder variant of $\mathcal{L}_{kd}$ and \textbf{only apply it on synthetic data} without freezing $\theta^{\mathcal{T}_{1:i}}$. We formulate our HKD as:
\begin{equation}
\mathcal{L}_{hkd} = \frac{1}{\lvert X^{old} \rvert \times \lvert \mathcal{T}_{1:i} \rvert} \sum_{\mathbf{x} \in X^{old}} \|
\theta_{i}\left(f_i\left( \mathbf{x} \right)\right) 
-
\theta^{\mathcal{T}_{1:i}}\left( 
f\left(\mathbf{x} \right)
\right)
\|_1.
\end{equation}

With this hard knowledge distillation, the outputs of old model $\theta_{i} \circ f_{i}$ and current model $\theta \circ f$ for the synthetic old data tend to be the same, but the model remains flexible inside to adapt to new knowledge. Next, we focus on learning representations of new classes, which requires the model to learn as many features from new task $\mathcal{T}_{i+1}$ as possible.

\smallskip\noindent\textbf{Local Classification Loss.} 
In CIL, it's common to use \textit{global} cross-entropy as the base loss that is applied on all available training data at the same time. However, when we use synthetic data to replace the real old data in DFCIL,  the domain gap between synthetic and real data leads the model to separate new and old classes by the difference of domain rather than semantics, as pointed out in ABD~\cite{abd}. We also observe that the decision boundaries within synthetic data 
are different from the ones within the real data. For instance, a synthetic fish image may mix a lot of features of a bird, which might confuse the old classifiers.
Therefore, we adopt a \textit{local} classification loss, which is the cross-entropy loss computed on the new data and the new classifiers (\ie, $\theta^{\mathcal{T}_{i+1}}$),  formulated as:
\begin{equation}
\mathcal{L}_{lce} = \frac{1}{|X^{new}|} \sum_{(\mathbf{x}, y) \in (X^{new}, Y^{new}) }
\mathcal{L}_{CE}\left(
\operatorname{softmax}\left(
\theta^{\mathcal{T}_{i+1}} \left( f \left( \mathbf{x} \right ) \right)
\right),
y
\right).
\end{equation}

This local classification loss does not directly affect classifiers of previous classes $\theta^{\mathcal{T}_{1:i}}$, but it changes $f$ to adapt to new task, which may corrupt the representations of previous learned classes. 

The conflict between learning representations of new classes and maintaining representations of previously learned classes can only be mitigated by sacrificing one for the other if the representation learning is not properly guided, and finally they compromise each other to achieve a coarse balance. 
Therefore, we propose to guide the current model to learn representations of new classes by the structural relation of their data in the old model's feature space. 

\smallskip\noindent\textbf{Relational Knowledge Distillation (RKD).} 
The HKD applied on synthetic data prevents changes in the representation of previous classes, since it strictly forces the representation of a single sample to be consistent on the new and old models. However, it  limits the model's plasticity when employed on data of new classes. In contrast to HKD, RKD~\cite{rkd} transfers structural information among a set of samples from teacher model to student model, endowing the student model more flexibility to learn new knowledge. The angle-wise RKD defines the relation on a triplet of samples $(\mathbf{x}_a, \mathbf{x}_b, \mathbf{x}_c)$ as the following cosine value:
\begin{equation}
\begin{aligned}
\text{cos} \angle \mathbf{r}_a\mathbf{r}_b\mathbf{r}_c  
= \left\langle\mathbf{e}^{a b}, \mathbf{e}^{cb}\right\rangle
\quad \text{where} \quad
\mathbf{e}^{i j} = \frac{\mathbf{r}_{i} - \mathbf{r}_{j}}{\left\|\mathbf{r}_{i}-\mathbf{r}_{j}\right\|_{2}}.
\end{aligned}
\end{equation}
Here $\mathbf{r}_{*}$ is sample $\mathbf{x}_{*}$'s feature representation on the teacher or student model. 

We incorporate RKD into our DFCIL framework, transferring the \textit{structural information of the new data} in the feature space of \textit{the old model} $\theta_i \circ f_i$ to current model $\theta \circ f$. Therefore, it can build a bridge between learning representations of new classes and maintaining representations of previously learned classes. 

When applying RKD, instead of directly using the data representations from the model to construct the relation, we first transform the representations  via  a $d \times 2d$ linear layer denoted as $\phi$, considering the following.  The new classes may not be effectively distinguished by their representations on old model $\theta_i \circ f_i$, and therefore, the structural relation built directly from the old representations may not help improving model's plasticity.
The representations of the new classes on current model $\theta \circ f$ are transformed by another linear layer $\psi$ to align to the transformed old representations. 
Then, we apply the following angle-wise relational knowledge distillation to a triplet $(\mathbf{x}_a, \mathbf{x}_b, \mathbf{x}_c)$ of the new data:
\begin{equation}
\begin{aligned}
&\mathcal{L}_{rkd} = \frac{1}{|X^{new}|^3}\sum_{\mathbf{x}_a, \mathbf{x}_b, \mathbf{x}_c \in X^{new}}\|
\cos \angle \mathbf{t}_a\mathbf{t}_b\mathbf{t}_c 
- 
\cos \angle \mathbf{s}_a\mathbf{s}_b\mathbf{s}_c
\|_1, \\
&\text{where} \quad 
\mathbf{t}_k = \phi\left(f_i\left(\mathbf{x}_k\right)\right), 
\mathbf{s}_k = \psi\left(f\left(\mathbf{x}_k\right)\right).
\end{aligned}
\end{equation}
\noindent By minimizing this loss, we limit the cases when the relation built from the transformed old representations hinders the improvement of plasticity or when the representation change hurts the model's stability. The two learnable transformation functions $\phi$ and $\psi$ are optimized with the representation learning to minimize this loss, making the relation distillation flexible. Using this relational knowledge distillation, we mitigate the conflict between improving plasticity by local classification loss and maintaining stability by hard knowledge distillation.

\smallskip\noindent\textbf{RRL Loss.} The above three components form the relation-guided representation learning (RRL). The loss of the RRL at phase $i$$+$$1$ is formulated as: 
\begin{equation}
\mathcal{L}_{rrl} = \lambda_{lce}^{i+1} \mathcal{L}_{lce} + \lambda_{hkd}^{i+1} \mathcal{L}_{hkd} + \lambda_{rkd}^{i+1} \mathcal{L}_{rkd},
\end{equation}
where lambdas are corresponding scale factors. Considering the amount of new knowledge increases with the number of new classes, and the difficulty of preserving previous knowledge grows as the ratio of previous classes to new classes gets larger, scale factors at learning phase $i$$+$$1$ are adaptively set as follows:
\begin{align}
&\lambda_{lce}^{i+1} = \frac{1 + 1 / \alpha}{\beta} \lambda_{lce}, \quad
\lambda_{hkd}^{i+1} = \alpha \beta \lambda_{hkd}, \quad
\lambda_{rkd}^{i+1} = \alpha \beta \lambda_{rkd} \\
&\text{where} \quad 
\alpha = \log_2(\frac{\lvert \mathcal{T}_{i+1} \rvert}{2} +1), \quad
\beta = ~\sqrt{\frac{\lvert \mathcal{T}_{1:i} \rvert}{\lvert \mathcal{T}_{i+1} \rvert}},
\end{align}
\noindent in which $\lambda_{lce}$, $\lambda_{hkd}$ and $\lambda_{rkd}$ are base scale factors that can be configurable, $\alpha$ and $\beta$ denote the amount of new knowledge and the difficulty of preserving previous knowledge, respectively. We appropriately increase the local classification loss to compensate for its weakening as the number of new classes decreases. The overall loss of the RRL at learning phase $i$$+$$1$ is finally defined as:
\begin{equation}
\mathcal{L}_{rrl} = \frac{1 + 1 / \alpha}{\beta} \lambda_{lce} \mathcal{L}_{lce} + \alpha \beta \lambda_{hkd} \mathcal{L}_{hkd} + \alpha \beta \lambda_{rkd} \mathcal{L}_{rkd}.
\end{equation}

\subsection{Classification Head Refinement} \label{sec:CHR} 
We achieve better stability-plasticity balance in feature extractor by relation-guided representation learning, but there are still two issues in classification head to address. One is that the decision boundaries between new and previous classes have not been learned by the model, and the other is that the imbalanced training data may cause biased classifiers. ABD attacks these problems concurrently with representation learning by a global task-balanced classification loss~\cite{dgr,ocfudw,bic}. However, we find that the global classification loss is not beneficial to the representation learning due to the domain gap between synthetic and real data. 
In addition to the data imbalance between new and previous classes, the data imbalance also exists within previous classes because the label of synthetic images are random. Inspired by prior works~\cite{podnet,bic}, we fine-tune the classification head with \textit{the feature extractor frozen} after representation learning, in which the $\mathcal{L}_{lce}$ is replaced with the following global class-balanced classification loss:  
\begin{equation}
\begin{aligned}
&\mathcal{L}_{gce} = \frac{1}{|X|} \sum_{(\mathbf{x}, y) \in (X, Y) }
\frac{w_y}{\sum_{j=0}^{\lvert \mathcal{T}_{1:i+1} \rvert - 1} w_j} \mathcal{L}_{CE}\left(
\operatorname{softmax}\left(
\theta \left(  f \left( \mathbf{x} \right ) \right)
\right),
y
\right), \\
&\text{where} \quad 
(X, Y) = (X^{new} \cup X^{old}, Y^{new} \cup Y^{old}).
\end{aligned}
\end{equation}
The weight $w_y$ of class $y$ is the reciprocal of it's number of samples (\ie, synthetic for previous classes and real for new classes) passed to the model during training.

\subsection{Image Synthesis}\label{sec:IS} 
The model inversion technique was first introduced to DFCIL in DeepInversion~\cite{deepinversion} to synthesize data for previous classes. 
DeepInversion iteratively optimizes random noises to images of given classes together with training the classification model, which is time consuming. 
Instead, ABD~\cite{abd} trains a synthesizer before learning new task, speeding up the learning process. 
Therefore, we follow ABD~\cite{abd} to train our synthesizer using the following four optimization objectives.
\textbf{The label diversity loss} forces the synthesizer to produce balanced data for previous classes.
\textbf{The data content loss} is the cross-entropy loss with a large temperature parameter to scale down the difference between the model's output, so that the synthetic images can be predicted as a certain class with high confidence. 
\textbf{The stat alignment loss} minimizes the KL divergence between the distribution of synthetic data and the distribution in BatchNorm layers of $f_i$, which record the statistics of the real data during the previous training.
\textbf{The image prior loss} encourages the synthesizer to produce more realistic images. 

By this means, we can obtain synthetic data that mimic the old real data. However, there are still the following issues with the synthetic data, and different techniques of our R-DFCIL addresses these issues accordingly. \textbf{1) Class imbalance} is attacked by class-balanced classification loss defined in Sec.~\ref{sec:CHR}. 
\textbf{2) The domain gap} between synthetic and real data, which misleads classifiers to learn wrong decision boundaries between new and previous classes, is attacked by separating the learning process into representation learning (Sec.~\ref{sec:RRL}) and classifier learning (Sec.~\ref{sec:CHR}).
\textbf{3) The conflict between model's plasticity and stability} is alleviated by relational knowledge distillation (Sec.~\ref{sec:RRL}), and catastrophic forgetting is effectively overcome by hard knowledge distillation. 
\section{Experiment}
\label{sec:experiment}

\subsection{Datasets and Evaluation Protocol} 

\noindent\textbf{Datasets.} We chose three representative classification datasets CIFAR100~\cite{cifar}, Tiny-ImageNet200~\cite{tiny_imagenet} and ImageNet100~\cite{ucir}, in which CIFAR100 and ImageNet100 are two extensively used datasets in CIL, and Tiny-ImageNet200 is considered as a challenging dataset for DFCIL~\cite{abd}. CIFAR100 contains 100 classes, each class with 500 training images of size 32$\times$32$\times$3 and 100 test images in the same size. ImageNet100 is a subset of ImageNet1000~\cite{imagenet}, with 100 randomly sampled classes. It has about 1300 training and 50 test images per class, and the spatial size of images vary. Tiny-ImageNet200 is an ImageNet-like dataset with smaller  (64$\times$64$\times$3) images than ImageNet. It has 200 classes in total, with 500 training and 50 test images for each class.

\noindent\textbf{Evaluation Protocol.} In the CIL literature, there are two commonly used protocols. The first protocol splits the classes equally into $N=5, 10, 20$ tasks for simulating short-term and large task incremental learning scenarios, in which $\lvert \mathcal{T}_{1:i} \rvert / \lvert \mathcal{T}_{i+1} \rvert$ is relatively small and the number of classes per task are relatively large. The other protocol introduced by Hou \etal~\cite{ucir} takes a half of classes as the first task, and equally divides the rest classes into 5, 10 or 25 tasks (\ie, $N=6, 11, 26$), which matches the situation of long-term and small task incremental learning. We follow prior works~\cite{icarl,ucir,podnet,abd} to evaluate approaches by the typical incremental metrics: last incremental accuracy $A_{N}$ and average incremental accuracy $\bar{A}_{N} = \frac{1}{N} \sum_{i=1}^N A_i$, in which the incremental accuracy $A_i$ is formally defined as:
\begin{equation}
A_i = \frac{1}{\lvert \mathcal{D}_{1:i}^{test} \rvert}
\sum_{(\mathbf{x}, y) \in \mathcal{D}_{1:i}^{test}} 
\mathds{1} \left(\hat{y} = y\right), 
~\text{where}~
\hat{y} = 
\mathop{\arg \max}_{0 \le j < \lvert \mathcal{T}_{1:i}\rvert}
\theta_{i}^{(j)}(f_i(\mathbf{x})),
\end{equation}
in which $\mathds{1}(\cdot)$ is the indicator function that maps the boolean value to $\{0, 1\}$.  

\setlength{\tabcolsep}{6.5pt}{
\renewcommand\arraystretch{0.7}
\begin{table}[tb]
  \caption{Evaluation on CIFAR100 with protocol that equally split 100 classes into $N$ tasks. The means and standard deviations are reported of three runs with random class orders. Approaches with * are reported directly from ABD paper.
  }
  \centering
  \begin{tabular}{lcccccc}
  \toprule
   \multirow{2.5}{*}{Approach}
   && $N=5$ && $10$ && $20$ \\
   \cmidrule{3-7}
   && $A_N$ (\%)  && $A_N$ (\%) && $A_N$ (\%) \\
   \midrule
   Upper Bound && 70.67 ± 0.16 && 70.67 ± 0.16 && 70.67 ± 0.16 \\ 
   \cmidrule{3-7}
   DGR*~\cite{dgr} && 14.40 ± 0.40 && 8.10 ± 0.10 && 4.10 ± 0.30 \\
   LwF*~\cite{lwf} && 17.00 ± 0.10 && 9.20 ± 0.00 && 4.70 ± 0.10 \\
   DeepInversion*~\cite{deepinversion} && 18.80 ± 0.30 && 10.90 ± 0.60 && 5.70 ± 0.30 \\
   ABD*~\cite{abd} && 43.90 ± 0.90 && 33.70 ± 1.20 && 20.00 ± 1.40 \\ 
   \cmidrule{3-7}
   ABD~\cite{abd} && 47.36 ± 0.48 && 36.19 ± 0.93 && 22.29 ± 0.65 \\
   \cmidrule{3-7}
   \textbf{R-DFCIL (Ours)} && \textbf{50.47 ± 0.43} && \textbf{42.37 ± 0.72} && \textbf{30.75 ± 0.12} \\
   \midrule
   && $\bar{A}_N$ (\%) && $\bar{A}_N$ (\%) && $\bar{A}_N$ (\%) \\
   \cmidrule{3-7}
   ABD~\cite{abd} &&  63.23 ± 1.49 && 56.61 ± 1.93  && 45.10 ± 2.01 \\
   \cmidrule{3-7}
   \textbf{R-DFCIL (Ours)} && \textbf{64.85 ± 1.78} && \textbf{59.41 ± 1.76} && \textbf{48.47 ± 1.90} \\
  \bottomrule
  \end{tabular}
  \label{tab:cifar100_p1}
\end{table}
}
\setlength{\tabcolsep}{2pt}{
\renewcommand\arraystretch{0.7}
\begin{table}[tb]
  \caption{
    Evaluation on CIFAR100 with the protocol introduced by Hou \etal~\cite{ucir}. The results of UCIR, PODNet and their Data-Free implementation UCIR-DF, PODNet-DF (all with CNN classifier) are present here for clearly comparison.
  }
  \centering
  \begin{tabular}{lcccccccc}
  \toprule
   \multirow{2.5}{*}{Approach} && \multirow{2.5}{*}{Data Free} && $N=6$ && $11$ && $26$ \\
   \cmidrule{4-9}
   &&&& $A_N$ (\%) && $A_N$ (\%) && $A_N$ (\%) \\
   \midrule
   UCIR (CNN)~\cite{ucir} && \ding{55} && 55.73 ± 0.89 && 53.22 ± 0.71 && 50.08 ± 0.35 \\
   PODNet (CNN)~\cite{podnet} && \ding{55} && 56.19 ± 1.00 && 52.53 ± 0.55 && 49.14 ± 0.25 \\
   \cmidrule{4-9}
   UCIR-DF (CNN)~\cite{ucir} && \ding{51} && 39.49 ± 0.81 && 25.54 ± 1.51 && 9.62 ± 0.73 \\
   PODNet-DF (CNN)~\cite{podnet} && \ding{51} && 40.54 ± 1.68 && 33.57 ±	2.48 && 20.18 ±	0.76 \\
   ABD~\cite{abd} && \ding{51} && 50.55 ± 1.14 && 43.65 ± 2.40 && 25.27 ± 1.09 \\
   \cmidrule{4-9}
   \textbf{R-DFCIL (Ours)} && \ding{51} && \textbf{54.76 ± 0.76} && \textbf{49.70 ± 0.61} && \textbf{30.01 ± 0.56} \\
   \midrule
   &&&& $\bar{A}_N$ (\%) && $\bar{A}_N$ (\%) && $\bar{A}_N$ (\%) \\
   \cmidrule{4-9}
   UCIR (CNN)~\cite{ucir} && \ding{55} && 65.58 ± 1.00 && 63.54 ± 1.12 && 60.32 ± 1.09 \\
   PODNet (CNN)~\cite{podnet} && \ding{55} && 66.82 ± 1.25 && 63.91 ± 1.07 && 61.56 ± 1.02 \\
   \cmidrule{4-9}
   UCIR-DF (CNN)~\cite{ucir} && \ding{51} && 57.82 ± 0.86 && 48.69 ± 1.16 && 33.33 ± 1.18 \\
   PODNet-DF (CNN)~\cite{podnet} && \ding{51} && 56.85 ± 1.40 && 52.61 ± 1.72 && 43.23 ± 1.70 \\
   ABD ~\cite{abd} && \ding{51} &&  62.40 ± 1.17 && 58.97 ± 1.87  && 48.91 ± 1.88 \\
   \cmidrule{4-9}
   \textbf{R-DFCIL (Ours)} && \ding{51} && \textbf{64.78 ± 1.58} && \textbf{61.71 ± 1.17} && \textbf{49.95 ± 0.76} \\
  \bottomrule
  \end{tabular}
  \label{tab:cifar100_p2}
\end{table}
}

\subsection{Implementation Details}
All approaches are implemented within the same code base written in PyTorch. We reproduce the current SOTA DFCIL approach ABD~\cite{abd}, and two popular replay-based CIL approaches UCIR~\cite{ucir}, PODNet~\cite{podnet}. To fairly comparision, we implement the UCIR-DF and PODNet-DF by replacing the real old data with the synthetic data that used by ABD and our R-DFCIL.
For CIFAR100, we follow the prior works~\cite{icarl,ucir,podnet} to adopt a modified 32-layer ResNet~\cite{resnet} backbone and train the model with SGD optimizer for 160 epochs, the learning rate is initially set to 0.1 and is divided by 10 after 80 and 120 epochs, the weight decay is set to 0.0005 and batch size is 128. We change the weight decay to 0.0002 for Tiny-ImageNet200 and keep other settings same as CIFAR100. For ImageNet100, we employ a ResNet18~\cite{resnet} backbone and train the model with SGD optimizer for 90 epochs, the learning rate starts from 0.1 and is divided by 10 after 30 and 60 epochs, the weight decay is set to 0.0001 and batch size is 64. Our R-DFCIL fine-tunes the classification head with a small constant learning rate 0.005 for another 40 epochs for CIFAR100, Tiny-ImageNet200, and 30 epochs for ImageNet100. The hyper parameters of our R-DFCIL are set to $\lambda_{lce}=0.5$, $\lambda_{hkd}=0.15$, $\lambda_{rkd}=0.5$ in all experiments. Please see supplementary material for more details on hyper-parameter tuning.

\setlength{\tabcolsep}{6.5pt}{
\renewcommand\arraystretch{0.7}
\begin{table}[tb]
  \caption{
    Evaluation on Tiny-ImageNet200 with the protocol that equally divides classes into $N$ tasks. The means and standard deviations are reported of three runs with random class orders. ABD* indicates data reported from ABD paper.
  }
  \centering
  \begin{tabular}{lcccccc}
  \toprule
   \multirow{2.5}{*}{Approach}
   && $N=5$ && $10$ && $20$ \\
   \cmidrule{3-7}
   && $A_N$ (\%) && $A_N$ (\%) && $A_N$ (\%) \\
   \midrule
   Upper Bound && 55.39 ± 0.33 && 55.39 ± 0.33 && 55.39 ± 0.33 \\
   \cmidrule{3-7}
   ABD*~\cite{abd} && - && -  && 12.1 \\
   \cmidrule{3-7}
   ABD~\cite{abd} && 30.56 ± 0.22 && 22.87 ± 0.67 && 15.20 ± 1.01 \\
   \cmidrule{3-7}
   \textbf{R-DFCIL (Ours)} && \textbf{35.89 ± 0.75} && \textbf{29.58 ± 0.51} && \textbf{24.43 ± 0.82} \\
   \midrule
   && $\bar{A}_N$ (\%) && $\bar{A}_N$ (\%) && $\bar{A}_N$ (\%) \\
   \cmidrule{3-7}
   ABD~\cite{abd} &&  45.30 ± 0.50 && 41.05 ± 0.54  && 34.74 ± 0.91 \\
   \cmidrule{3-7}
   \textbf{R-DFCIL (Ours)} && \textbf{48.96 ± 0.40} && \textbf{44.36 ± 0.18} && \textbf{39.34 ± 0.18} \\
  \bottomrule
  \end{tabular}
  \label{tab:tiny200_p1}
\end{table}
}
\setlength{\tabcolsep}{6.5pt}{
\renewcommand\arraystretch{0.8}
\begin{table}[tb]
  \caption{
    Evaluation on Tiny-ImageNet200 with the protocol introduced by Hou \etal~\cite{ucir}. The means and standard deviations are reported of three runs with random class orders. The best values are in bold font.
  }
  \centering
  \begin{tabular}{lcccccc}
  \toprule
   \multirow{2.5}{*}{Approach}
   && $N=6$ && $11$ && $26$ \\
   \cmidrule{3-7}
   && $A_N$ (\%) && $A_N$ (\%) && $A_N$ (\%) \\
   \midrule
   ABD~\cite{abd} && 33.18 ± 0.46 && 27.34 ± 0.44 && 16.46 ± 0.34 \\
   \cmidrule{3-7}
   \textbf{R-DFCIL (Ours)} && \textbf{40.44 ± 0.11} && \textbf{38.19 ± 0.08} && \textbf{27.29 ± 0.24} \\
   \midrule
   && $\bar{A}_N$ (\%) && $\bar{A}_N$ (\%) && $\bar{A}_N$ (\%) \\
   \cmidrule{3-7}
   ABD~\cite{abd} &&  44.55 ± 0.13 && 41.64 ± 0.46 && 34.47 ± 0.29 \\
   \cmidrule{3-7}
   \textbf{R-DFCIL (Ours)} && \textbf{48.91 ± 0.29} && \textbf{47.60 ± 0.50} && \textbf{40.85 ± 0.28} \\
  \bottomrule
  \end{tabular}
  \label{tab:tiny200_p2}
\end{table}
}

\subsection{Results and Analysis}

\smallskip\noindent\textbf{CIFAR100.} We follow ABD~\cite{abd} to conduct five-, ten-, and twenty-tasks class-incremental experiments, with respectively 20, 10 and 5 classes per task. We run all approaches on three random class orders with the seeds 0, 1, 2 (\ie, consistent with the official ABD code) and report the means and standard deviations of these three runs. In Table~\ref{tab:cifar100_p1}, we report the results of ABD implemented by us and present the original data reported by ABD paper. Our R-DFCIL surpasses ABD by $3.11$$/$$1.62$ ($A_N / \bar{A}_N$), $6.18$$/$$2.80$ and $8.46$$/$$3.37$ percent points on five-, ten-, and twenty-tasks settings, respectively. Table~\ref{tab:cifar100_p2} shows results of the experiments with the protocol introduced by Hou \etal~\cite{ucir}, in which the first task has 50 classes and 10, 5, 2 classes per incremental task for $N=6, 11, 26$, respectively. From the comparison between UCIR/PODNet and UCIR-DF/PODNet-DF, we can see a great performance degradation of the popular replay-based approaches when replacing the real old data with synthetic old data. Prior CIL works believe that more tasks imply stronger forgetting. But we find that the initially learned knowledge and the number of classes in incremental tasks also impact forgetting, since both ABD and our R-DFCIL perform better with the second protocol than with the first protocol despite more tasks (Table~\ref{tab:cifar100_p2} \vs ~\ref{tab:cifar100_p1}). 

\begin{figure*}[tb]
  \centering
  \begin{subfigure}[b]{0.49\linewidth}
    \includegraphics[width=\linewidth]{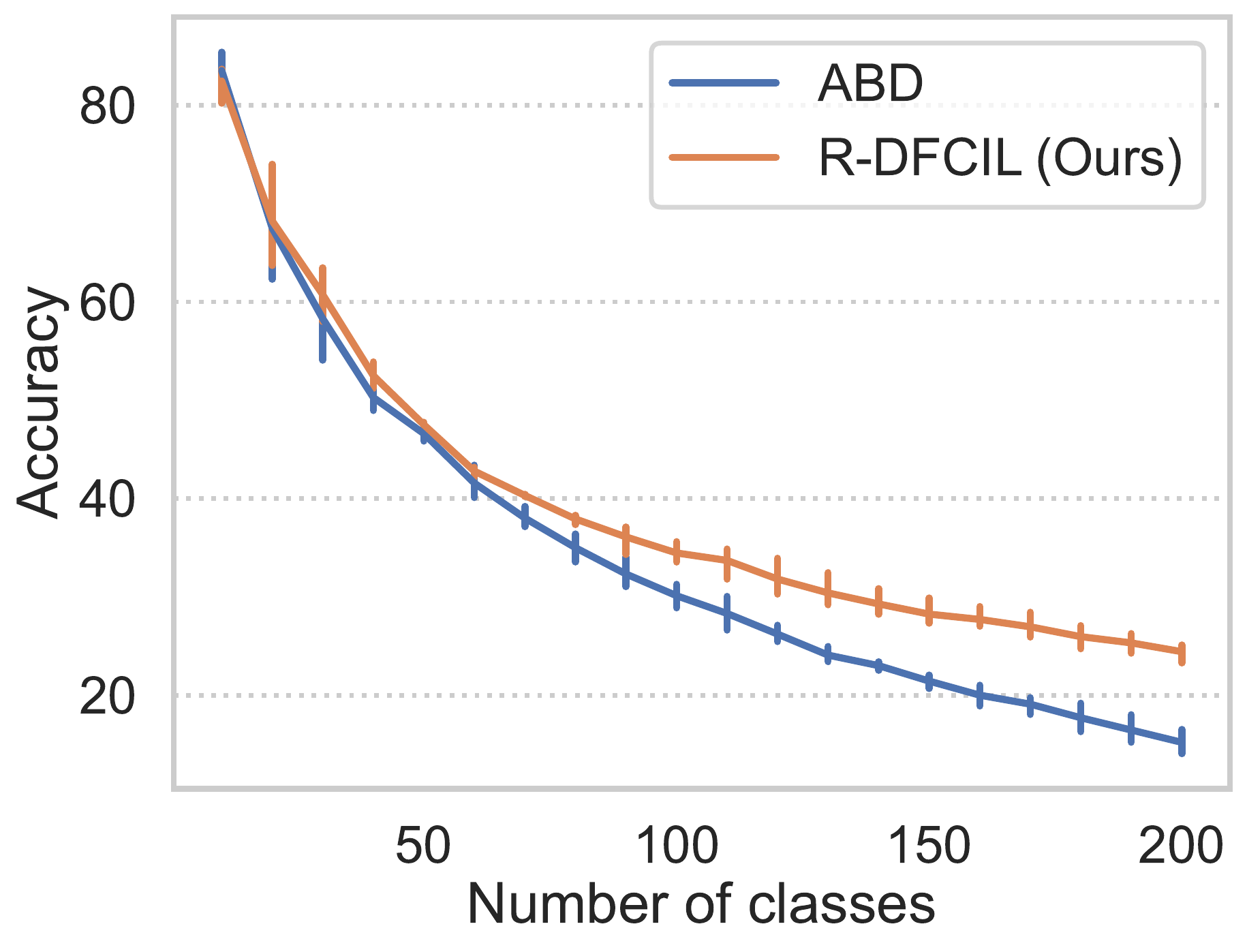}
    \caption{20 tasks, 10 classes / incremental task}
    \label{fig:cifar_inc1}
  \end{subfigure}
  \hfill
  \begin{subfigure}[b]{0.49\linewidth}
    \includegraphics[width=\linewidth]{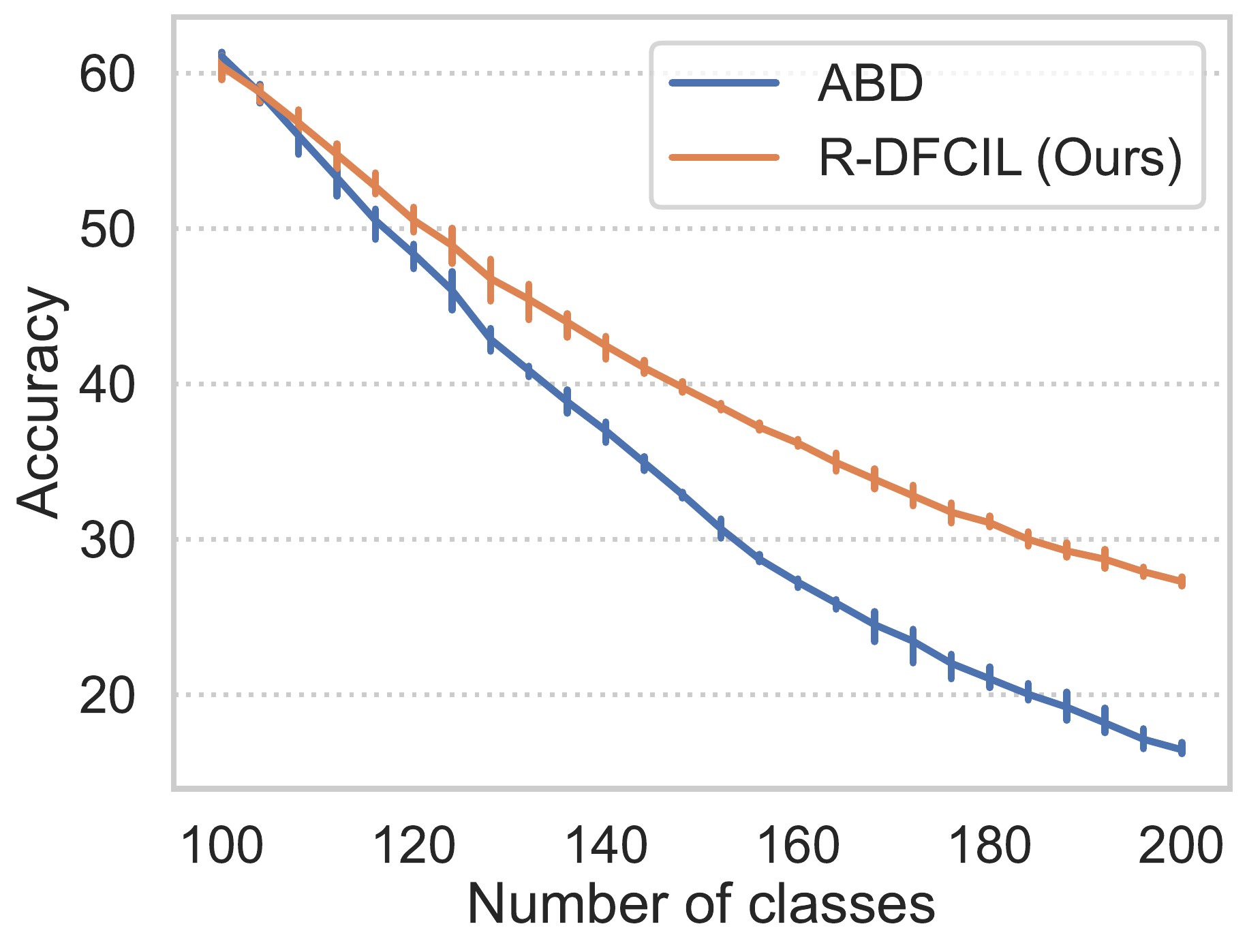}
    \caption{26 tasks, 4 classes / incremental task}
    \label{fig:cifar_inc2}
  \end{subfigure}
  \caption{\textbf{Incremental Accuracy on Tiny-ImageNet200}. The lines show the phase-by-phase evaluation results of ABD~\cite{abd} and our F-DFCIL. The means and standard deviations are reported of three runs with random class orders.}
  \label{fig:tiny_inc_acc}
\end{figure*}

\smallskip\noindent\textbf{Tiny-ImageNet200.}
We compare our R-DFCIL with ABD in the more challenging dataset Tiny-ImageNet200, in which we can observe similar results to the experiments on CIFAR100. From the data presented in Table~\ref{tab:tiny200_p1}, \ref{tab:tiny200_p2}, we can see that there are more performance gains of our R-DFCIL over ABD as the total number of tasks increases (\eg, $N = 5 \to 20$, $6 \to 26$). We plot the task-by-task incremental accuracy in Fig. \ref{fig:tiny_inc_acc}, in which we can see the ABD drops faster than our R-DFCIL as the number of learned classes increases. We can conclude from the above observations that our R-DFCIL solves the forgetting of previously learned classes better than ABD.

\setlength{\tabcolsep}{12pt}{
\renewcommand\arraystretch{0.7}
\begin{table}[tb]
  \caption{
    Evaluation on ImageNet100 with the protocol that equally split 100 classes into $N$ tasks. We report the evaluation results of a single run.
  }
  \centering
  \begin{tabular}{lcccccc}
  \toprule
   \multirow{2.5}{*}{Approach}
   && $N=5$ && $10$ && $20$ \\
   \cmidrule{3-7}
   && $A_N$ (\%) && $A_N$ (\%) && $A_N$ (\%) \\
   \midrule
   Upper Bound && 77.46 && 77.46 && 77.46 \\
   \cmidrule{3-7}
   ABD~\cite{abd} && 51.46 && 35.96 && 22.40  \\
   \cmidrule{3-7}
   \textbf{R-DFCIL (Ours)} && \textbf{53.10} && \textbf{42.28} && \textbf{30.28} \\     	  	 	 
   \midrule
   && $\bar{A}_N$ (\%) && $\bar{A}_N$ (\%) && $\bar{A}_N$ (\%) \\
   \cmidrule{3-7}
   ABD~\cite{abd} && 67.42 && 57.76  && 44.89  \\
   \cmidrule{3-7}  	 
   \textbf{R-DFCIL (Ours)} && \textbf{68.15} && \textbf{59.10} && \textbf{47.33} \\
  \bottomrule
  \end{tabular}
  \label{tab:imagenet100_p1}
\end{table}
}

\smallskip\noindent\textbf{ImageNet100.}
We report the experimental results on ImageNet100 in Table~\ref{tab:imagenet100_p1}. In these experiments, the model is less prone to forgetting than experiments on CIFAR100 and Tiny-ImageNet200 due to the large model capacity (11.0 \vs 0.4 million parameters). Although the performance of ABD is close to our R-DFCIL when $N$$=$$5$, the difference becomes significant when $N$ increases to 20.

\begin{figure}[tb]
  \centering
  \begin{subfigure}[b]{0.49\linewidth}
    \includegraphics[width=\linewidth]{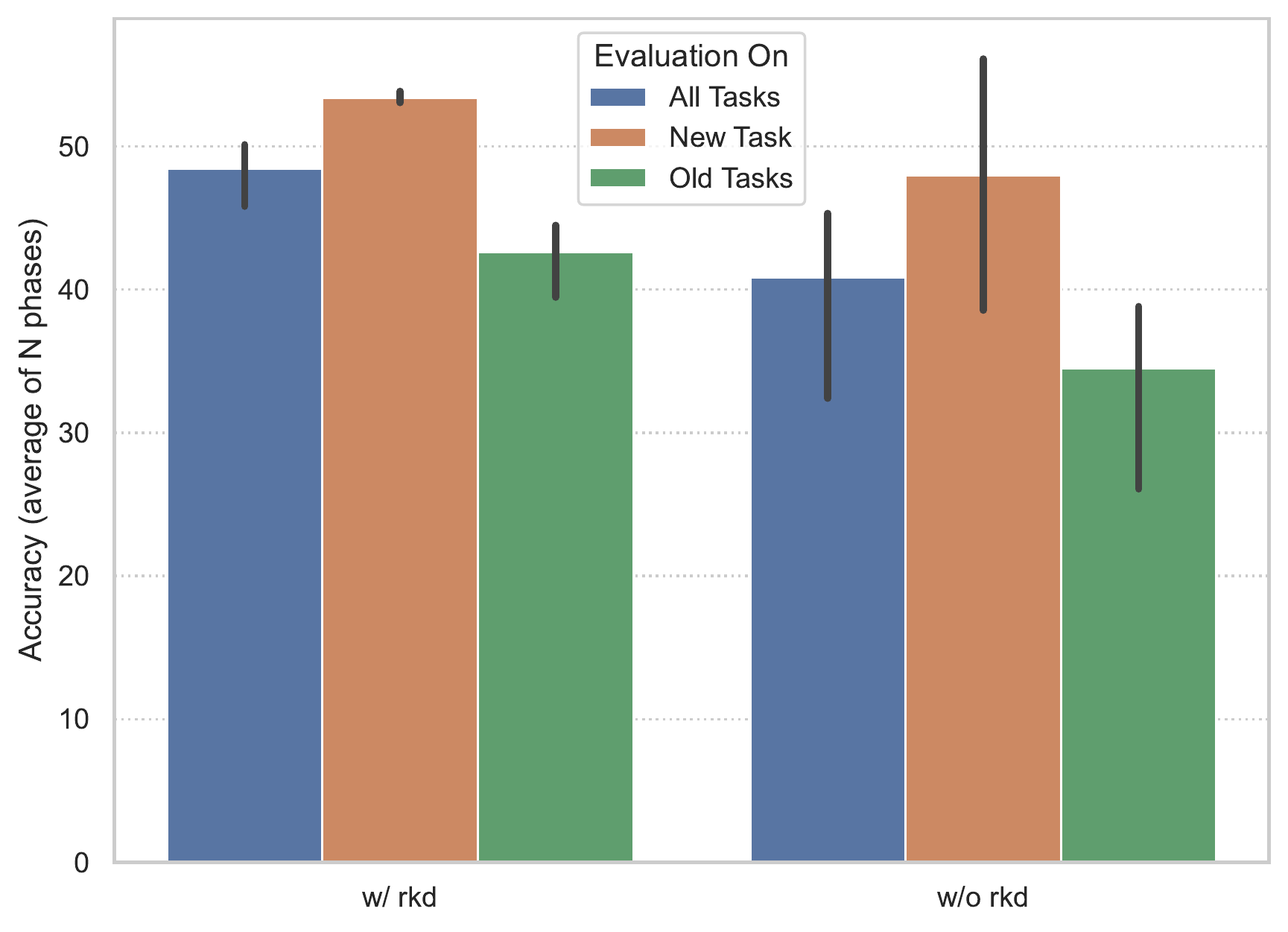}
    \caption{relational knowledge distillation}
  \end{subfigure}
  \hfill
  \begin{subfigure}[b]{0.49\linewidth}
    \includegraphics[width=\linewidth]{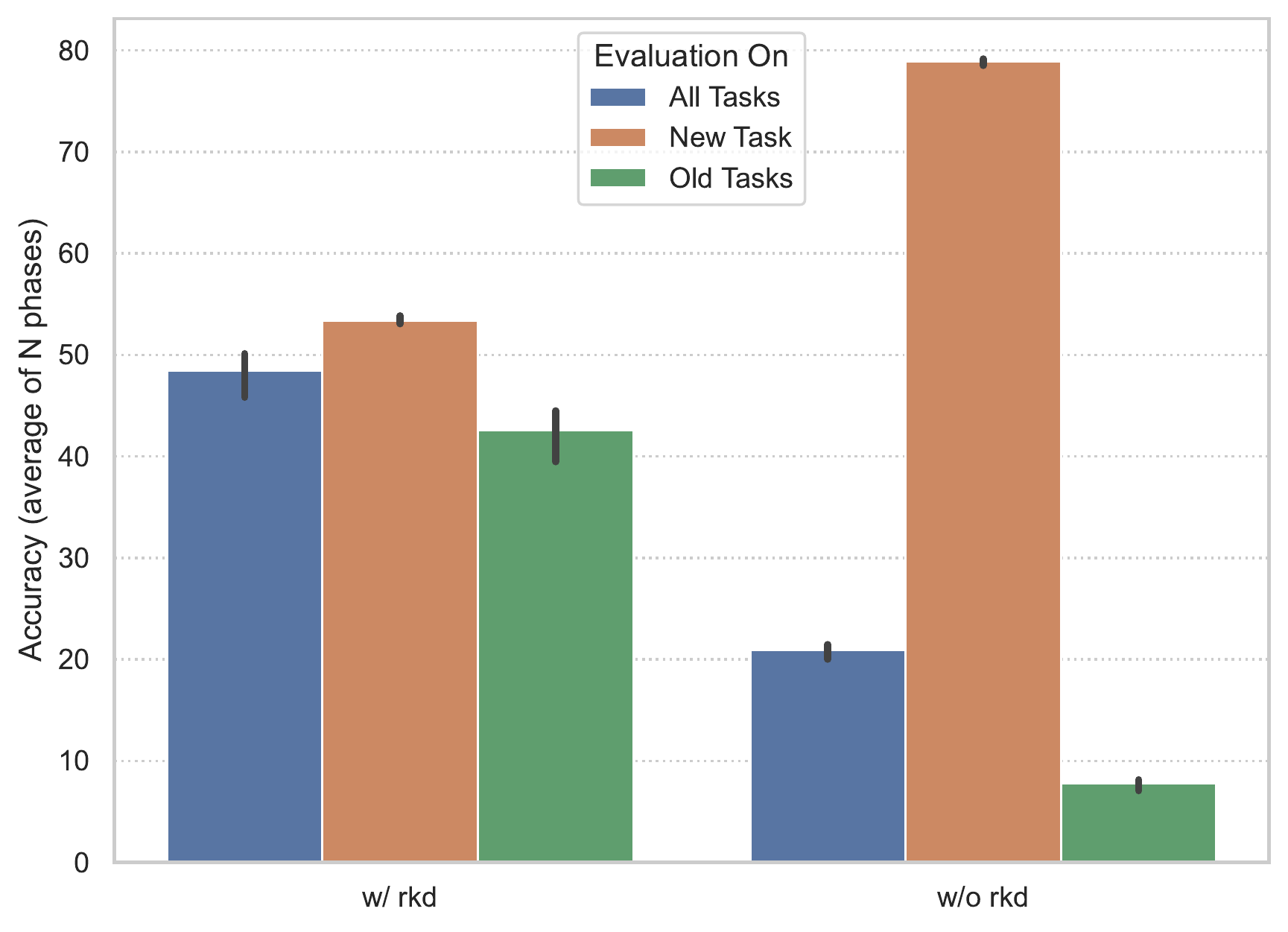}
    \caption{hard knowledge distillation}
  \end{subfigure}
  \caption{\textbf{Ablation Study about Stability-Plasticity Balance}. The left (a) shows a better balance with RKD (w/ rkd), and the right show the importance of the HKD to mitigate forgetting.}
  \label{fig:ablation}
  \vspace{-15pt}
\end{figure}

\smallskip\noindent\textbf{Ablation Study.}
We ablate three main components of our R-DFCIL, and display the results in Table~\ref{tab:ablation}. The experiments are conducted on CIFAR100 with total $N$$=$$20$ tasks and $5$ classes per task. All three components contribute greatly to our R-DFCIL, the last incremental accuracy drops by 9.21, 25.38, 7.00 percent point without relational knowledge distillation (RKD), hard knowledge distillation (HKD), classification head refinement (CHR), respectively. From Fig.~\ref{fig:ablation}, we can clearly see that the HKD is necessary for reducing the forgetting of learned classes. We also observe that the RKD boost both plasticity and stability, demonstrating the success of our relation-guided representation learning in alleviating the conflict between improving plasticity and maintaining stability. In fact, our R-DFCIL achieves better plasticity as well as stability than previous approaches, the details are present in supplementary material. It is worth emphasizing that our adaptive design (\ie, introduction of linear transformation functions) contributes about 2\% gain in the last incremental accuracy. We also investigated some newer relational KD methods, please see supplementary material.

\setlength{\tabcolsep}{9.2pt}{
\renewcommand\arraystretch{0.8}
\begin{table}[tb]
  \caption{
    Abalation Study on CIFAR100 with $N=20$. The results show the comparison between our R-DFCIL with all components and without relation knowledege distillation (RKD), hard knowledge distillation (HKD), classification head refinement (CHR, \ie, training process ends with representation learning).
  }
  \centering
  \begin{tabular}{lcccccccc}
  \toprule
   RKD && HKD && CHR && $A_N$ (\%) && $\bar{A}_N$ (\%) \\
   \midrule
   \ding{55} && \ding{51} &&  \ding{51} && 21.63 ± 5.60 && 40.86 ± 5.98 \\
   \cmidrule{7-9}
   \ding{51} && \ding{55} &&  \ding{51} &&  5.37 ± 0.35 && 20.96 ± 0.69 \\
   \cmidrule{7-9}
   \ding{51} && \ding{51} &&  \ding{55} && 23.75 ± 0.81 && 43.09 ± 1.53 \\
   \cmidrule{7-9}
   \ding{51} && \ding{51} &&  \ding{51} && \textbf{30.75 ± 0.12} && \textbf{48.47 ± 1.90} \\
  \bottomrule
  \end{tabular}
  \label{tab:ablation}
  \vspace{-15pt}
\end{table}
}
\section{Conclusion}
\label{sec:conclusion}

This paper studies the problem of Data-Free Class-Incremental Learning (DFCIL). We propose relation-guided representation learning (RRL) for DFCIL (R-DFCIL) to address the catastrophic forgetting caused by the severe domain gap between synthetic and real data. In RRL, the model overcomes forgetting of previous classes by hard knowledge distillation on synthetic data, and learns new knowledge by the local classification loss on new data. The relational knowledge distillation (RKD) can mitigate the conflict between improving plasticity and maintaining stability by transferring structural relation of new data from the old to the current model. After RRL, the classification head is refined with global class-balanced classification loss to address data imbalance issue and learn the decision boundaries between classes. Our R-DFCIL surpasses previous SOTA approach on CIFAR100, Tiny-ImageNet200 and ImageNet100 with 8.46\%, 9.23\%, and 9.88\% accuracy gain, respectively. Our R-DFCIL learns representation and classifier independently in two stages, which constructs a basic framework for future studies to address the domain gap between synthetic and real data in DFCIL. We introduce RKD to DFCIL for the first time, providing a reference for future works to overcome forgetting using structural information.

\noindent \textbf{Acknowledgement} This work was supported by the King Abdullah University of Science and Technology (KAUST) Office of Sponsored Research (OSR) under Award No. OSR-CRG2021-4648 and the Shenzhen General Research Project (JCYJ20190808182805919).

\bibliographystyle{splncs04}
\bibliography{egbib}

\appendix
\newpage
\addcontentsline{toc}{section}{Supp}
\section*{
\centering\Large
Supplementary Material
}

In the supplementary materials, we further validate the proposed approach by providing the following:
\begin{itemize}
\item Section~\ref{sec:rkd}: Relational Knowledge Distillation in Detail.
\item Section~\ref{sec:asf}: Details of Adaptive Scale Factors in RRL Loss.
\item Section~\ref{sec:exp}: Additional Experimental Details.
\item Section~\ref{sec:exp_res}: Additional Experimental Results.
\item Section~\ref{sec:fa}: Feature Analysis and Bottlenecks in Prior Approaches.
\end{itemize}

\section{Relational Knowledge Distillation in Detail}
\label{sec:rkd}

\begin{figure}
\centering
\includegraphics{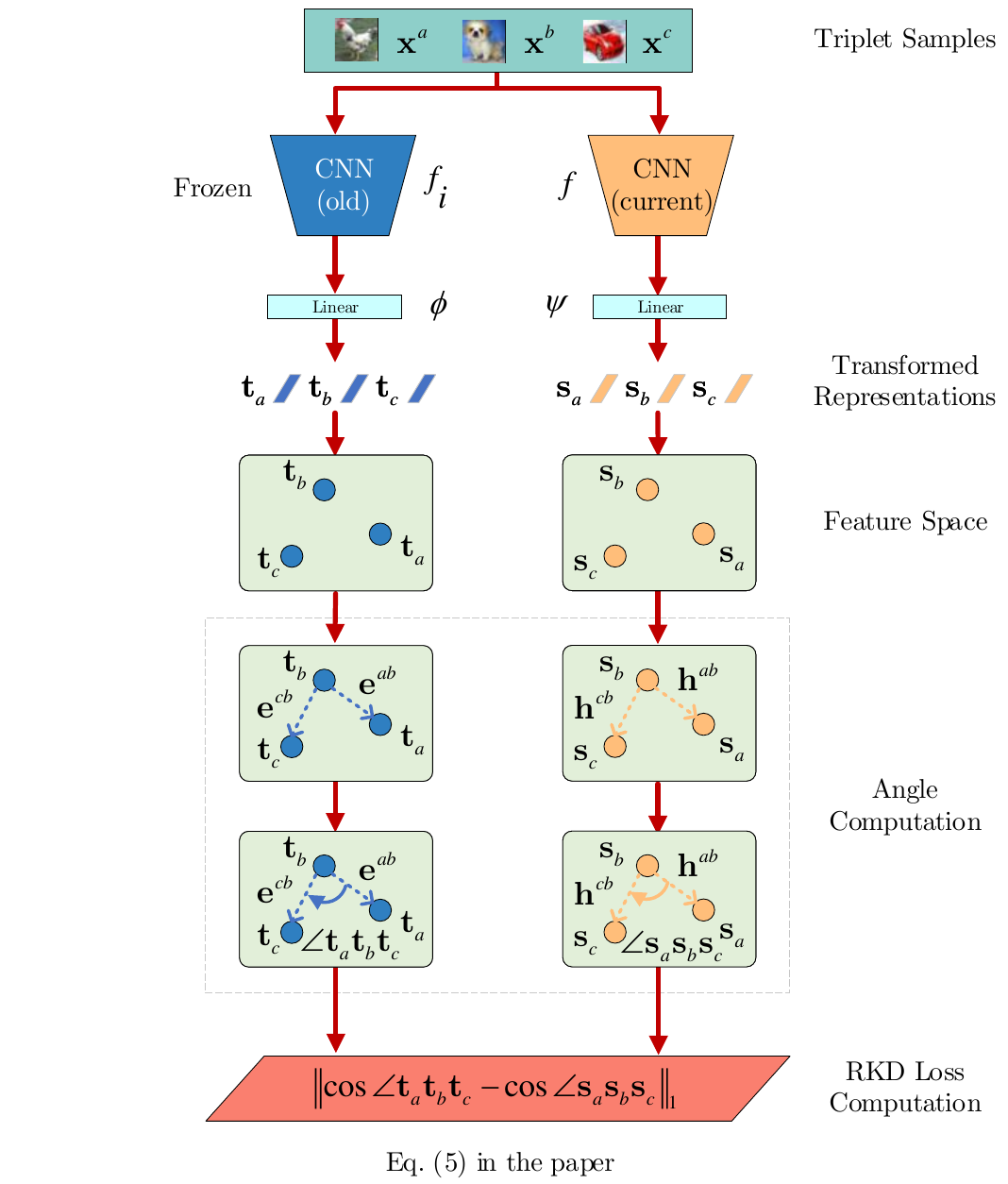}
\caption{\textbf{Relational Knowledge Distillation in Detail.} We first extract samples' representations, then transform them by linear layers. The angle is computed in the feature space and RKD loss is calculated by Eq. (5) in the paper.}
\label{fig:rkd}
\end{figure}

We described the relation knowledge distillation (RKD) in Sec. 3.2 of the main paper. Here we give some more details of RKD, by illustrating the process 
in Fig.~\ref{fig:rkd}. We calculate the RKD loss for a triplet of new samples ($\mathbf{x}_a$,$\mathbf{x}_b$,$\mathbf{x}_c$) in several steps. 
\textbf{First}, we extract their representations on the old model ($f_i$) and the current model ($f$), respectively. 
\textbf{Then}, we  transform the representations by two linear layers $\phi$ and $\psi$, and we get two groups of transformed representations $\mathbf{t}_{*}$ and $\mathbf{s}_{*}$, all of which are points ($\mathbb{R}^{2d}$ vectors) in the feature space. 
\textbf{Next}, from the point $\mathbf{t}^b$  we construct edges $\mathbf{e}^{ab}=\mathbf{t}^a-\mathbf{t}^b$ and $\mathbf{e}^{cb}=\mathbf{t}^c-\mathbf{t}^b$ in the feature space, and obtain the angle $\angle \mathbf{t}_a \mathbf{t}_b \mathbf{t}_c$ between the two edges. Similarly, we can get the angle $\angle \mathbf{s}_a \mathbf{s}_b \mathbf{s}_c$ by constructing edges $\mathbf{h}^{ab}=\mathbf{s}^a-\mathbf{s}^b$ and $\mathbf{h}^{cb}=\mathbf{s}^c-\mathbf{s}^b$. 
\textbf{Finally}, we compute the RKD loss about the  sample $\mathbf{x}_b$ by Eq. (5) in the main paper. The RKD loss about the other two samples $\mathbf{x}_a$ and $\mathbf{x}_c$ are computed in the same way, which are omitted in the figure.

\section{Details of Adaptive Scale Factors in RRL Loss}
\label{sec:asf}

In  Sec. 3.2 of the main paper, we elaborated our relation-guide representation learning (RRL), in which hard knowledge distillation (HKD) prevents  forgetting of previous knowledge, local cross-entropy loss (LCE) improves the model's plasticity, and  relational knowledge distillation (RKD) alleviates the conflict between them. However, the scale of these three components should adapt to the situations in practice since the number of classes the model has learned, and the number of classes in the new task vary at different times. \textbf{For one thing}, the larger the number of previous classes is compared to that of new classes $\frac{\lvert \mathcal{T}_{1:i} \rvert}{\lvert \mathcal{T}_{i+1} \rvert}$, the relatively more previous knowledge the model has to maintain, \ie, the more difficult it is to prevent forgetting. Therefore, in Eq. (9) of the main paper, we scaled up the losses for HKD and RKD $\mathcal{L}_{hkd}$, $\mathcal{L}_{rkd}$ and scaled down the loss for LCE $\mathcal{L}_{lce}$ by $\beta = \sqrt{\frac{\lvert \mathcal{T}_{1:i} \rvert}{\lvert \mathcal{T}_{i+1} \rvert}}$. \textbf{For another}, the effect of $\mathcal{L}_{lce}$ becomes stronger as the number of new classes grows, which also increases the difficulty of preserving previous knowledge. For this reason, we scaled up $\mathcal{L}_{hkd}$, $\mathcal{L}_{rkd}$ by $\alpha = \log_2(\frac{\lvert \mathcal{T}_{i+1} \rvert}{2} +1)$, which was carefully selected from a group of functions that are positively correlated to $\lvert \mathcal{T}_{i+1}\rvert \ge 2$ and start from 1 (when $\lvert \mathcal{T}_{i+1} \rvert$$=$$2$). In addition, the LCE loss $\mathcal{L}_{lce}$ gets weak when the number of classes is very small due to the reduction in the number of negative classes (\ie, $\lvert \mathcal{T}_{i+1} \rvert - 1$), so we compensate the $\mathcal{L}_{lce}$ by $\frac{1}{\alpha}$. This adaptive strategy makes our RRL work better in various complex incremental learning situations.

\begin{figure}
  \centering
  \begin{subfigure}{\linewidth}
    \centering
    \begin{subfigure}{0.49\linewidth}
      \includegraphics[width=\linewidth]{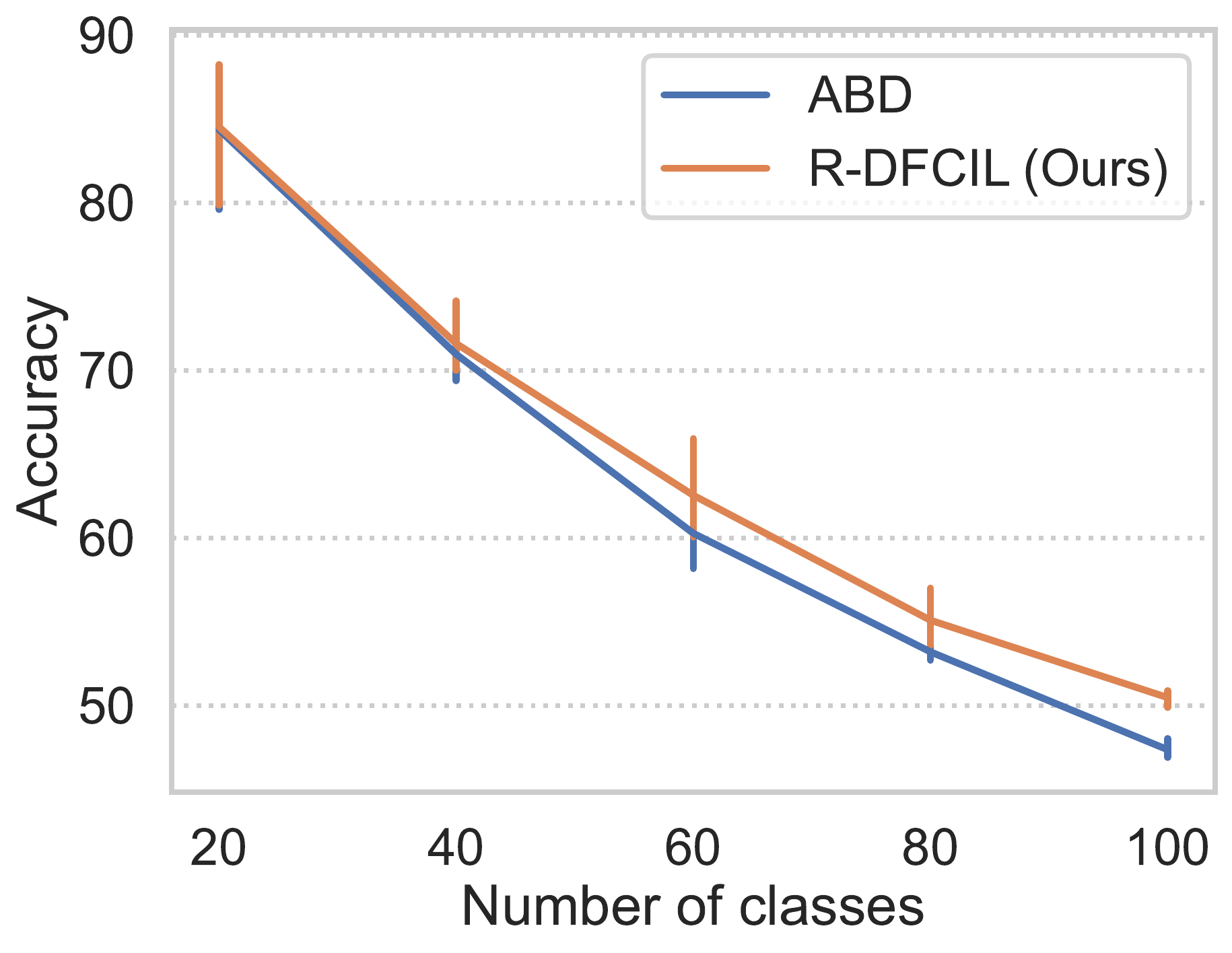}
      \caption{5 tasks, 20 classes / incremental task}
      \label{fig:cifar_inc_1_1}
    \end{subfigure}
    \begin{subfigure}{0.49\linewidth}
      \includegraphics[width=\linewidth]{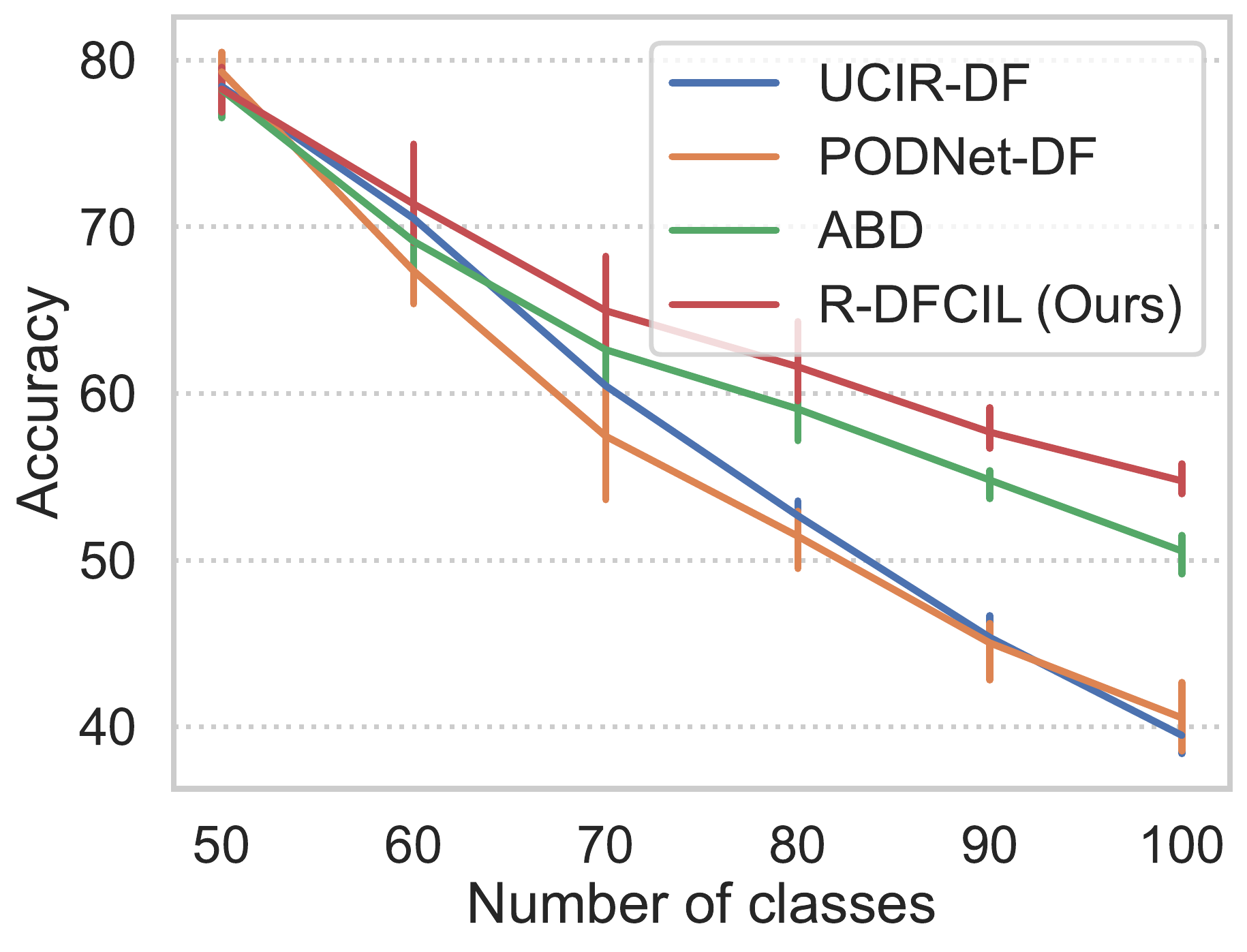}
      \caption{6 tasks, 10 classes / incremental task}
      \label{fig:cifar_inc_1_2}
    \end{subfigure}
  \end{subfigure}
  \begin{subfigure}{\linewidth}
    \centering
    \begin{subfigure}{0.49\linewidth}
      \includegraphics[width=\linewidth]{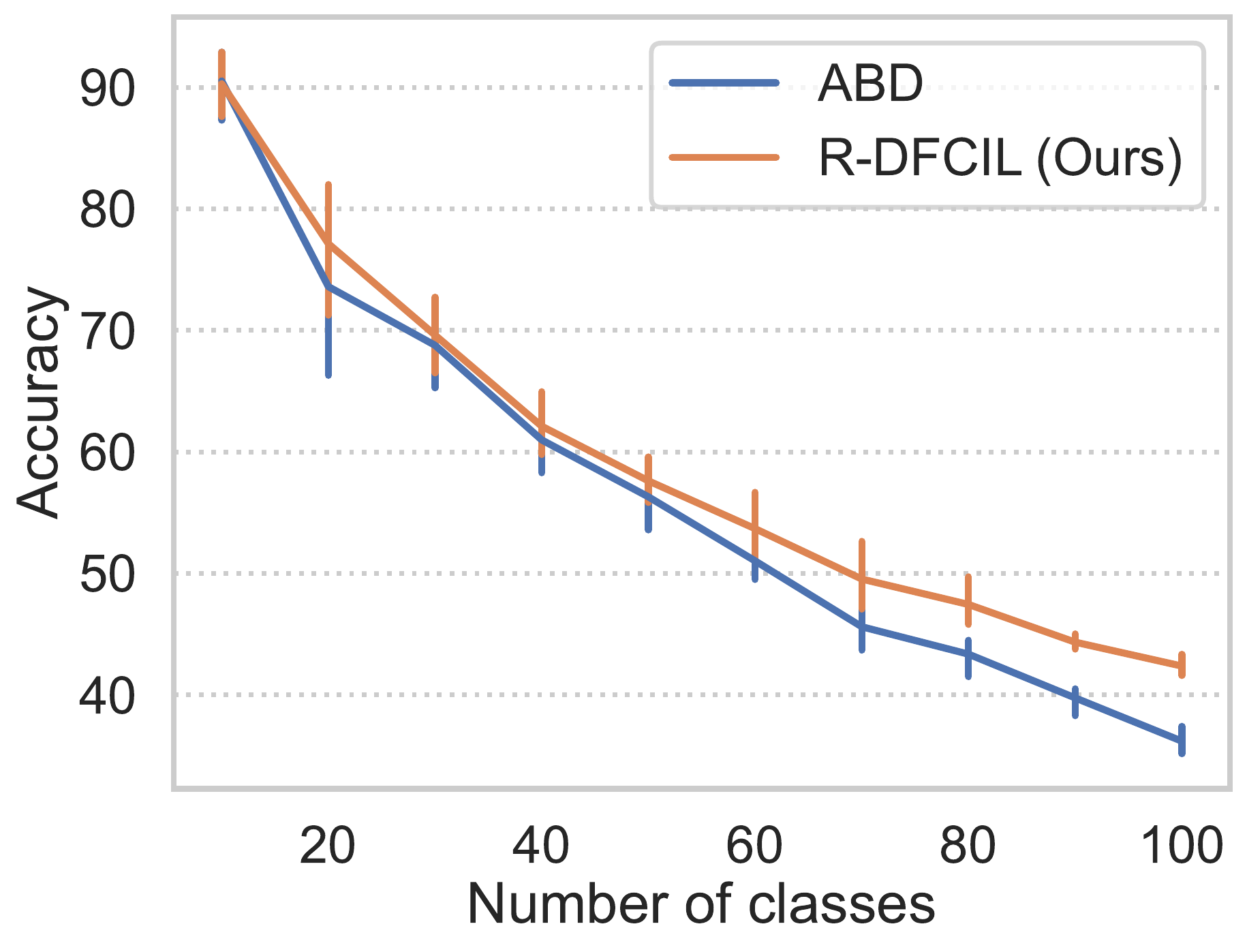}
      \caption{10 tasks, 10 classes / incremental task}
      \label{fig:cifar_inc_2_1}
    \end{subfigure}
    \begin{subfigure}{0.49\linewidth}
      \includegraphics[width=\linewidth]{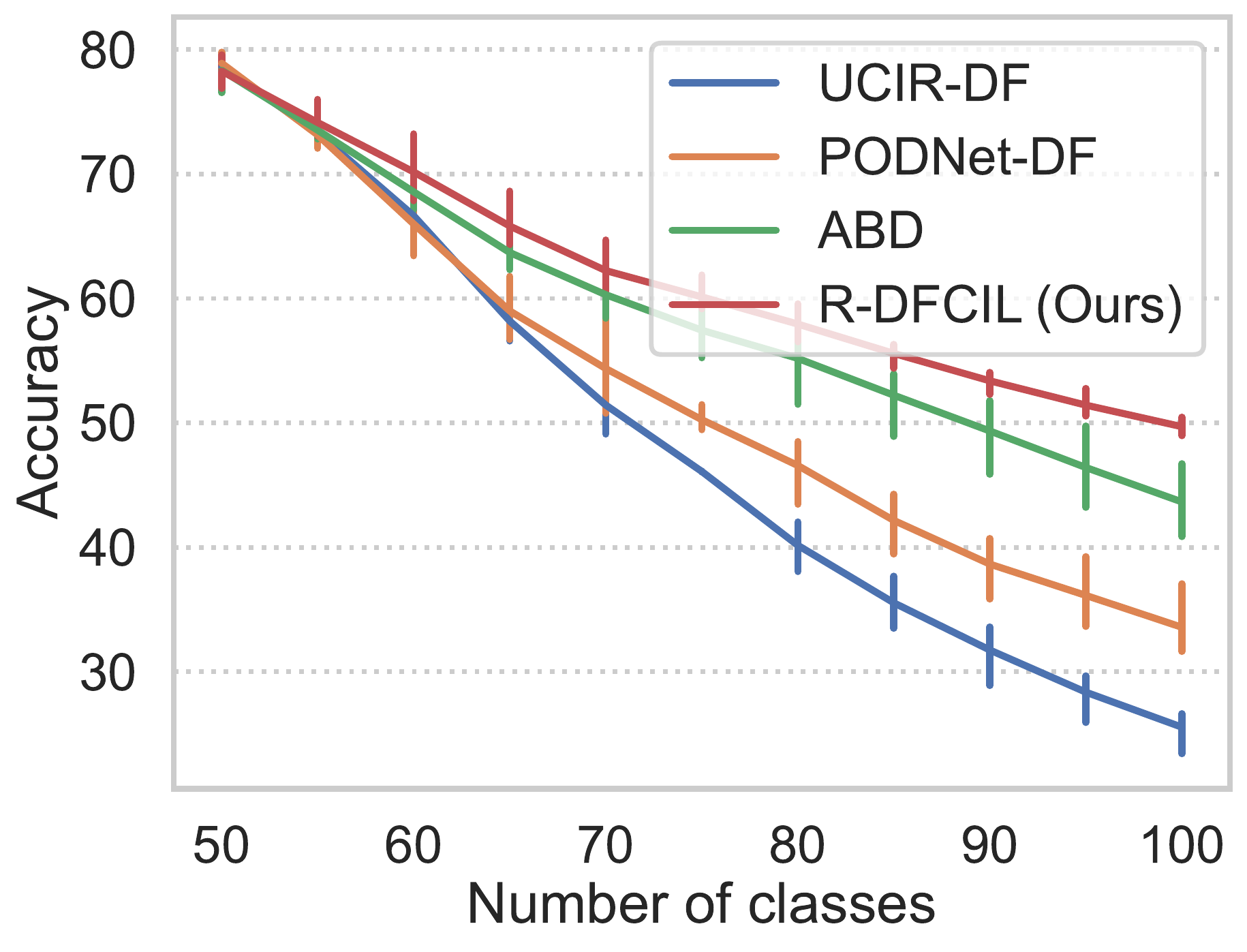}
      \caption{11 tasks, 5 classes / incremental task}
      \label{fig:cifar_inc_2_2}
    \end{subfigure}
  \end{subfigure}
  \begin{subfigure}{\linewidth}
    \centering
    \begin{subfigure}{0.49\linewidth}
      \includegraphics[width=\linewidth]{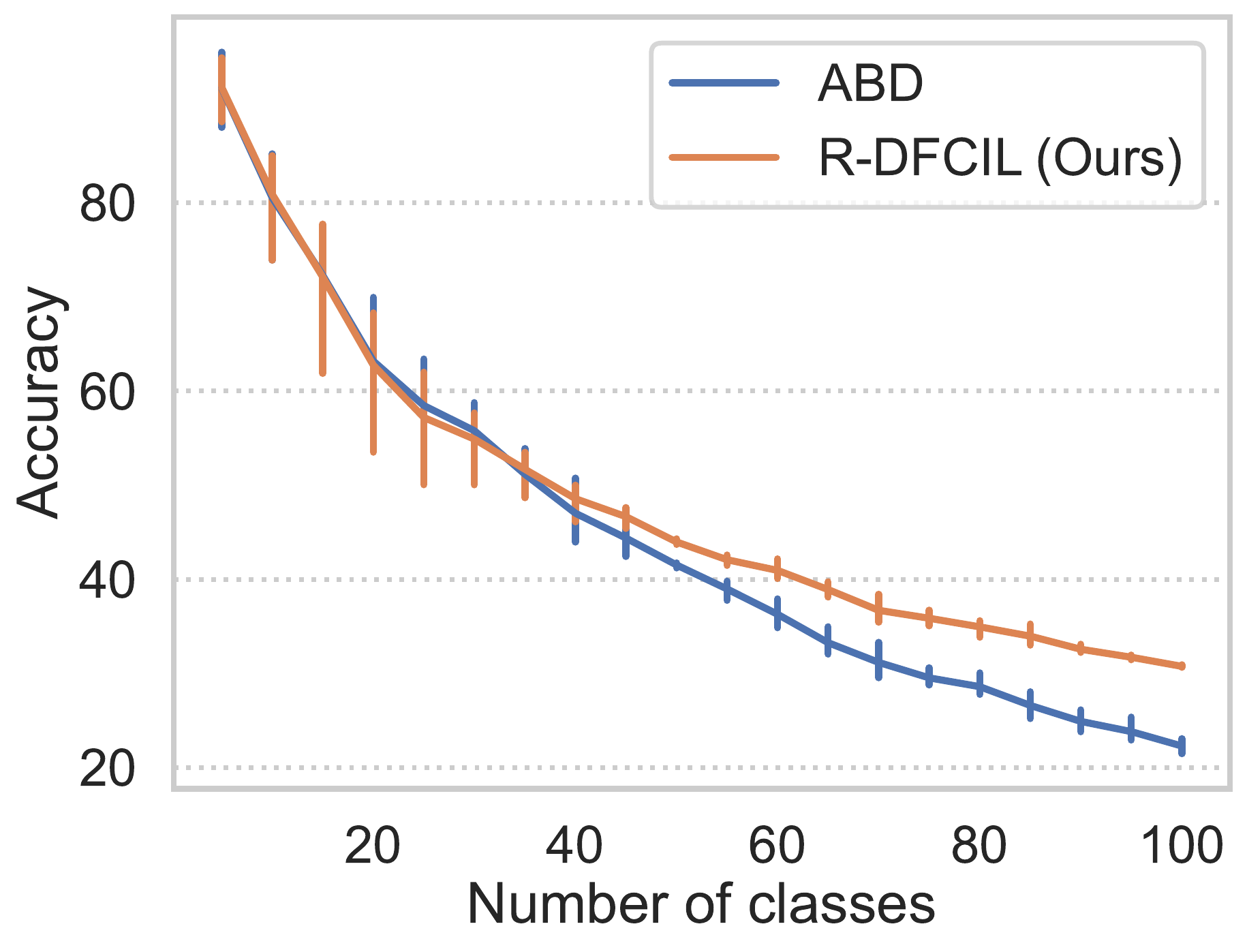}
      \caption{20 tasks, 5 classes / incremental task}
      \label{fig:cifar_inc_3_1}
    \end{subfigure}
    \begin{subfigure}{0.49\linewidth}
      \includegraphics[width=\linewidth]{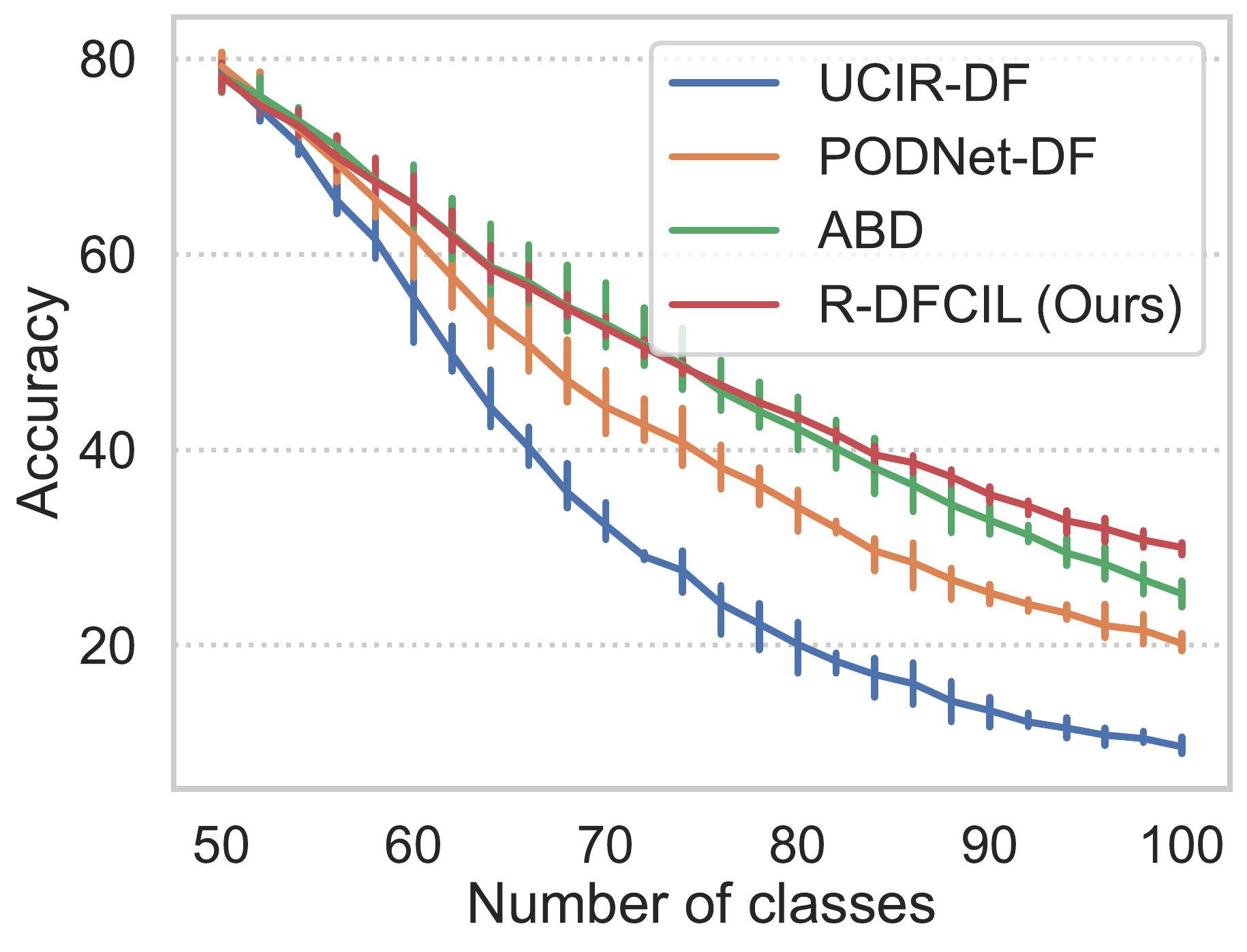}
      \caption{26 tasks, 2 classes / incremental task}
      \label{fig:cifar_inc_3_2}
    \end{subfigure}
  \end{subfigure}
  \caption{\textbf{Incremental Accuracy on CIFAR100}. 
  The margins between our R-DFCIL and other approaches gradually increase as the number of learned classes grows. The popular CIL approaches (UCIR, PODNet) work badly with synthetic data (UCIR-DF, PODNet-DF). The means and standard deviations are reported of three runs with random class orders.
  }
  \label{fig:cifar_inc_acc}
\end{figure}
\begin{figure}
  \centering
  \begin{subfigure}{\linewidth}
    \centering
    \begin{subfigure}{0.49\linewidth}
      \includegraphics[width=\linewidth]{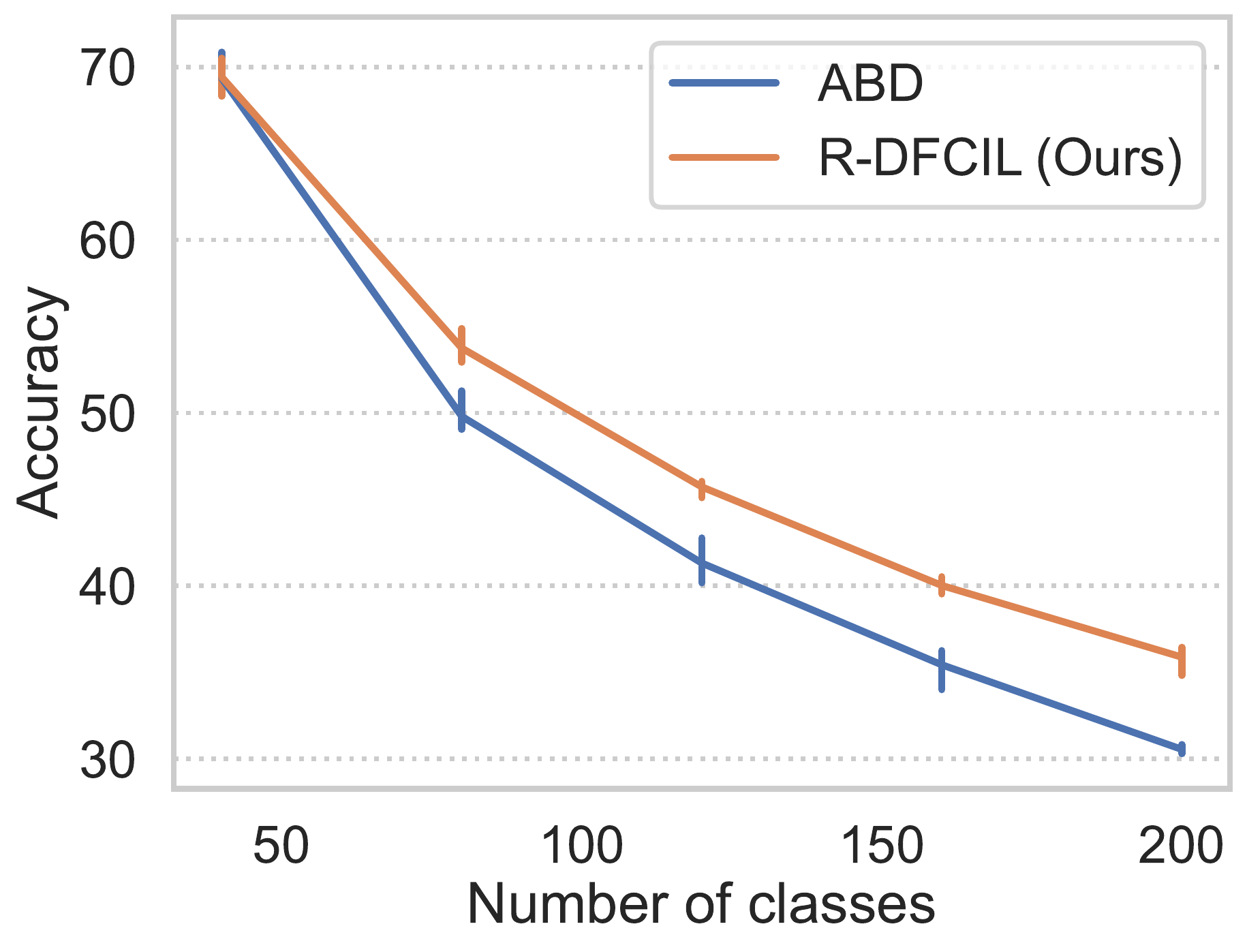}
      \caption{5 tasks, 40 classes / incremental task}
    \end{subfigure}
    \begin{subfigure}{0.49\linewidth}
      \includegraphics[width=\linewidth]{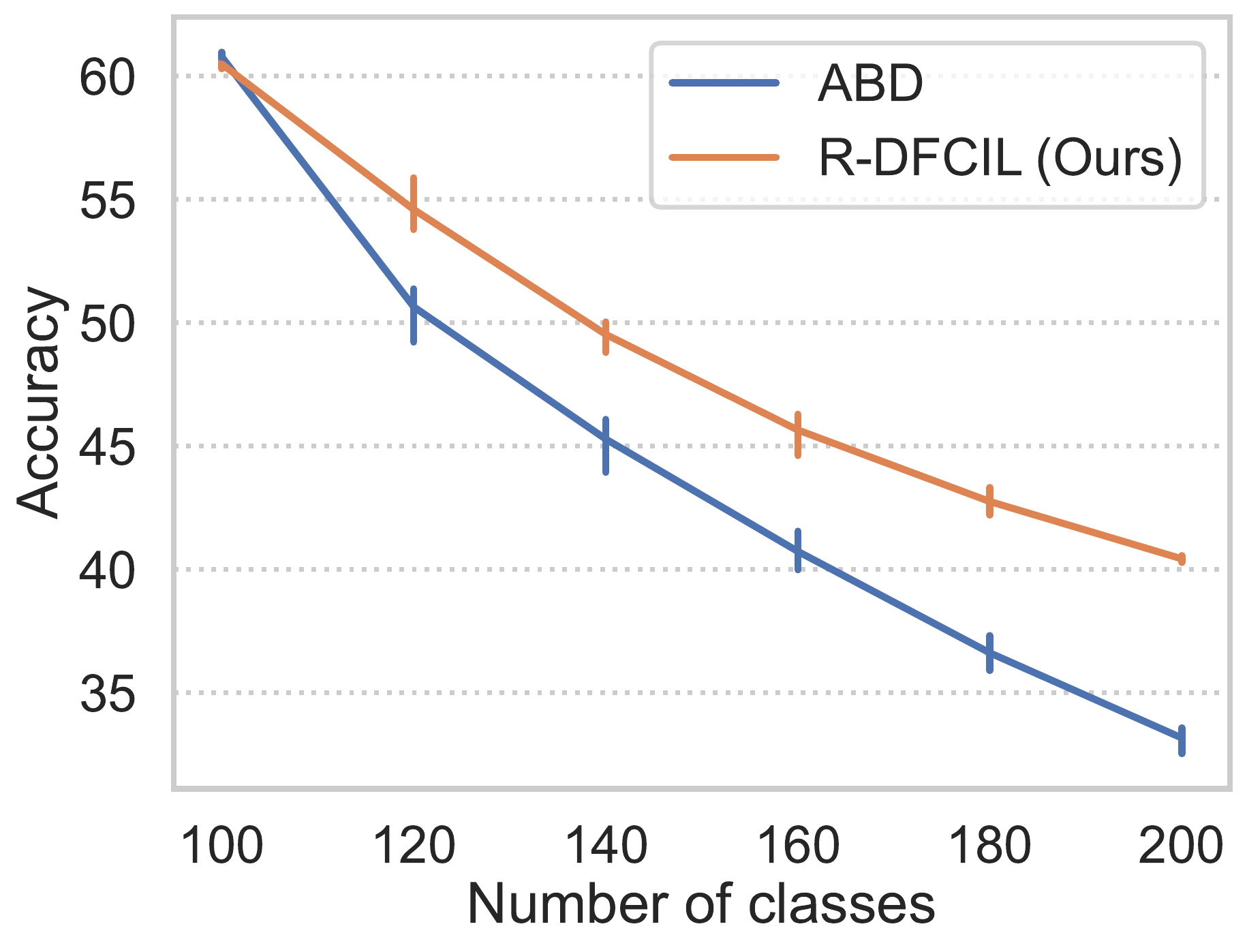}
      \caption{6 tasks, 20 classes / incremental task}
    \end{subfigure}
  \end{subfigure}
  \begin{subfigure}{\linewidth}
    \centering
    \begin{subfigure}{0.49\linewidth}
      \includegraphics[width=\linewidth]{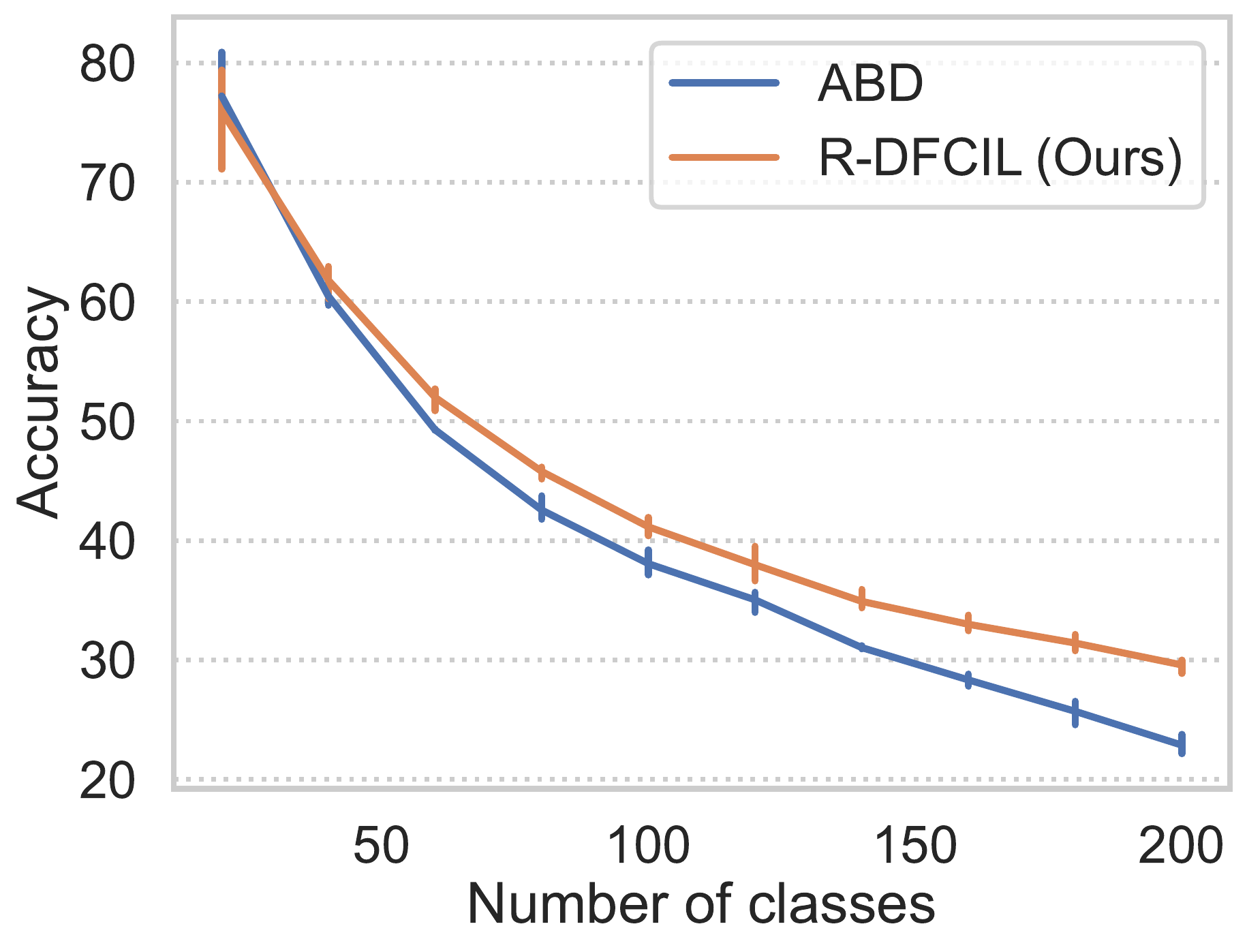}
      \caption{10 tasks, 20 classes / incremental task}
    \end{subfigure}
    \begin{subfigure}{0.49\linewidth}
      \includegraphics[width=\linewidth]{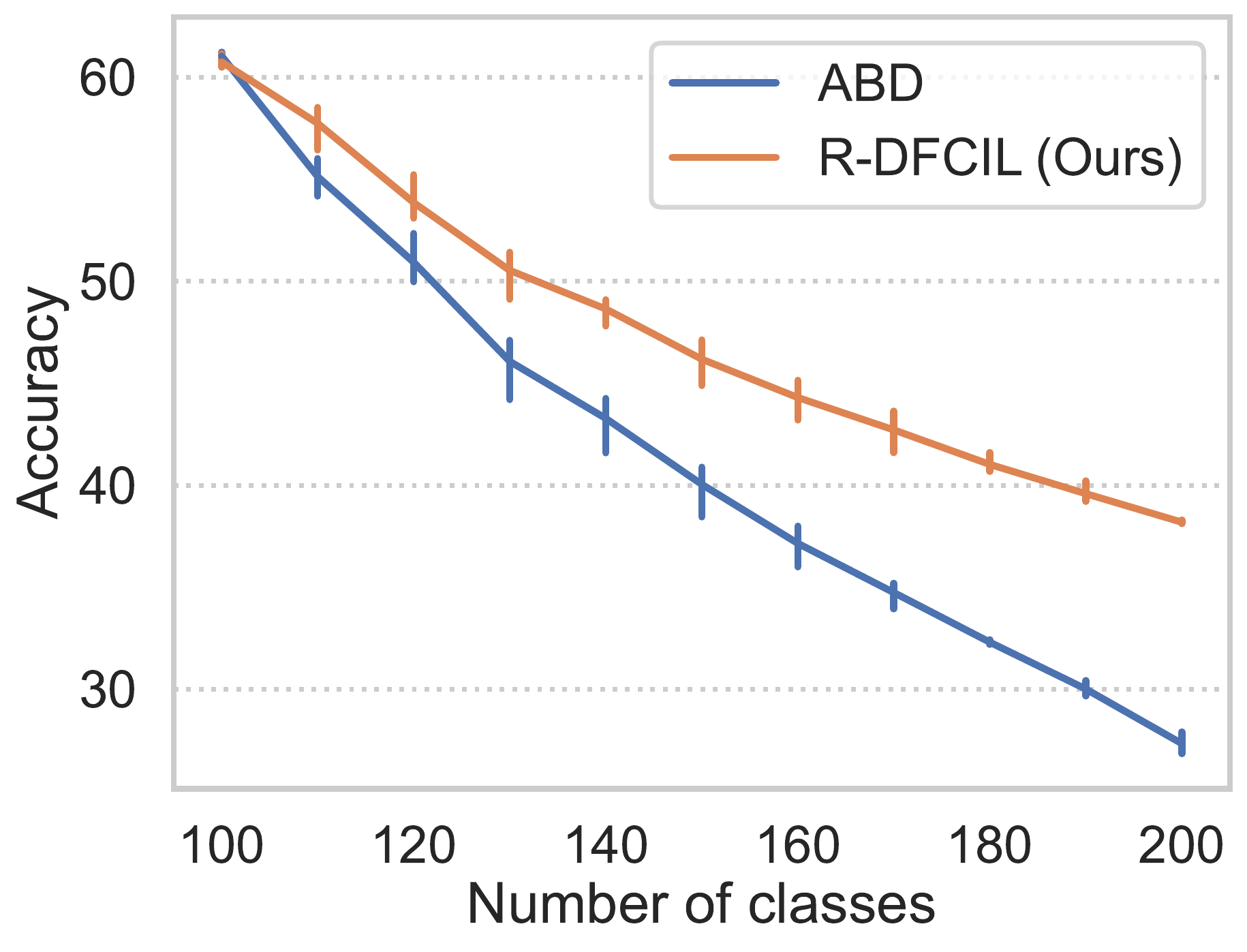}
      \caption{11 tasks, 10 classes / incremental task}
    \end{subfigure}
  \end{subfigure}
  \caption{\textbf{Incremental Accuracy on Tiny-ImageNet200}.
  The differences between our R-DFCIL and ABD are more sigificant on Tiny-ImageNet200 than on CIFAR100, though Tiny-ImageNet200 is more challenging. This may because the model is more prone to forgetting when there are more classes in incremental task, and our R-DFCIL is better at mitigating forgetting. The means and standard deviations are reported of three runs with random class orders.
  }
  \label{fig:tiny_inc_acc_sup}
\end{figure}
\begin{figure}
  \centering
  \vspace{-10pt}
  \begin{subfigure}[b]{0.24\linewidth}
    \includegraphics[width=\linewidth]{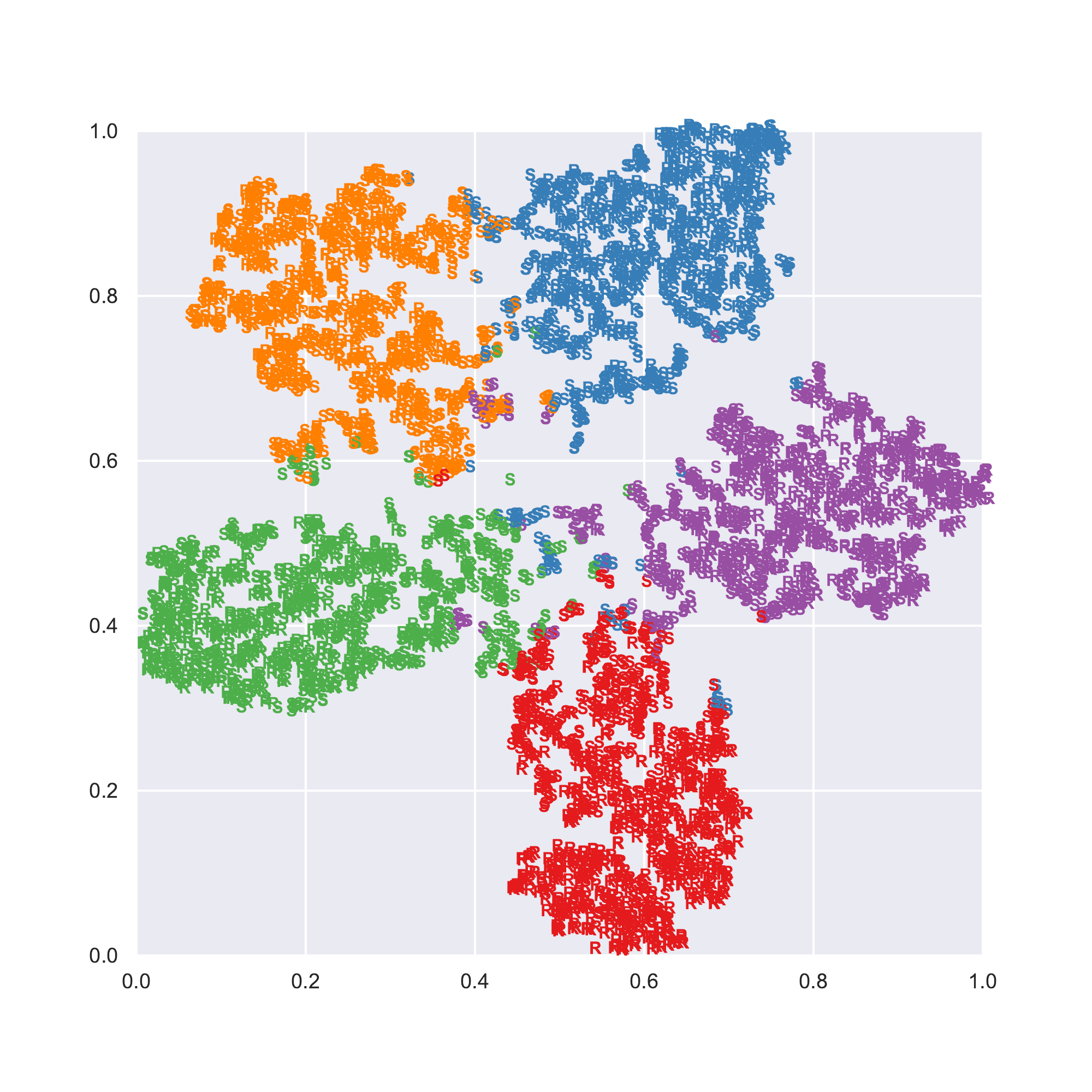}
    \caption{Joint}
    \label{fig:tsne_joint}
  \end{subfigure}
  \hfill
  \begin{subfigure}[b]{0.24\linewidth}
    \includegraphics[width=\linewidth]{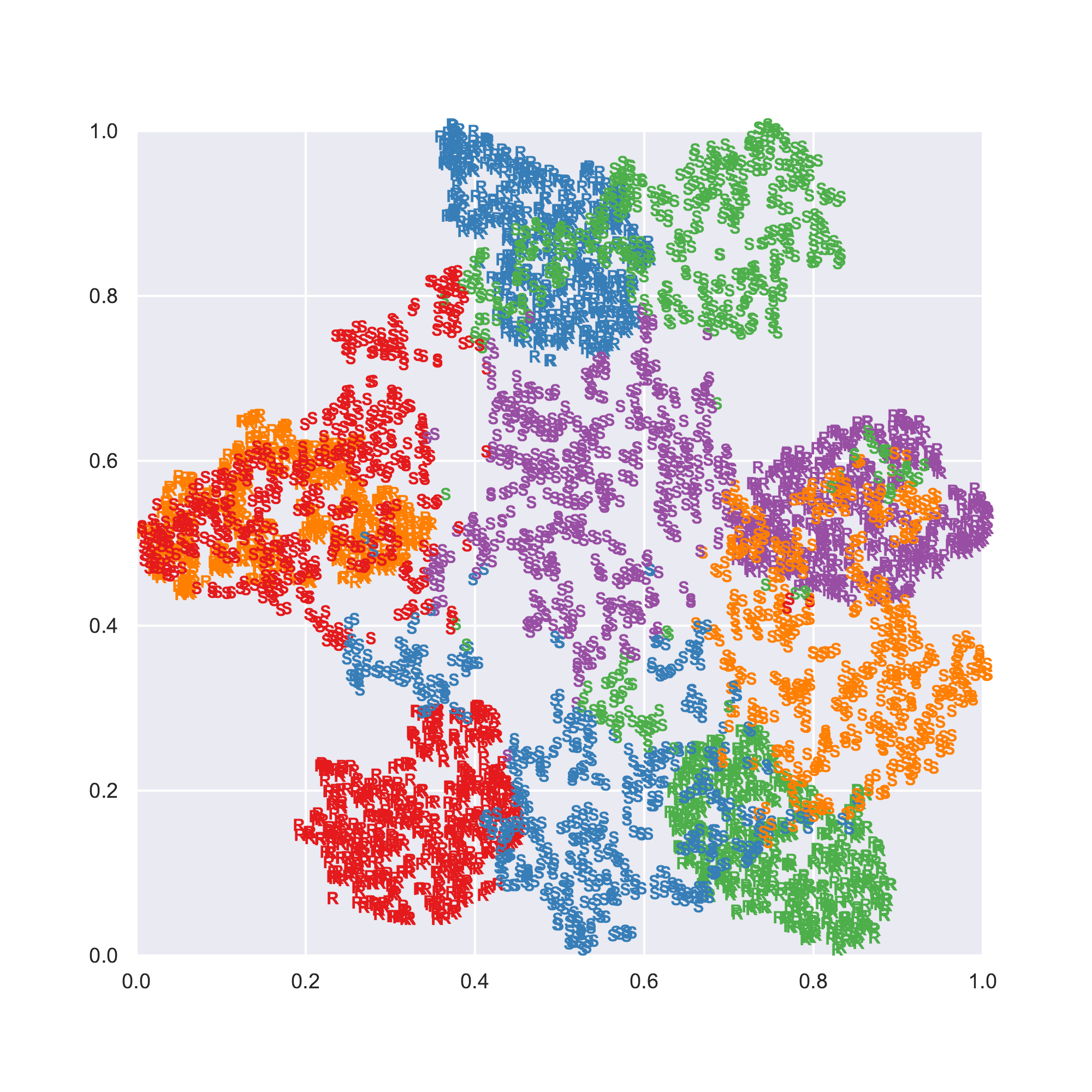}
    \caption{Independent}
    \label{fig:tsne_independent}
  \end{subfigure}
  \begin{subfigure}[b]{0.24\linewidth}
    \includegraphics[width=\linewidth]{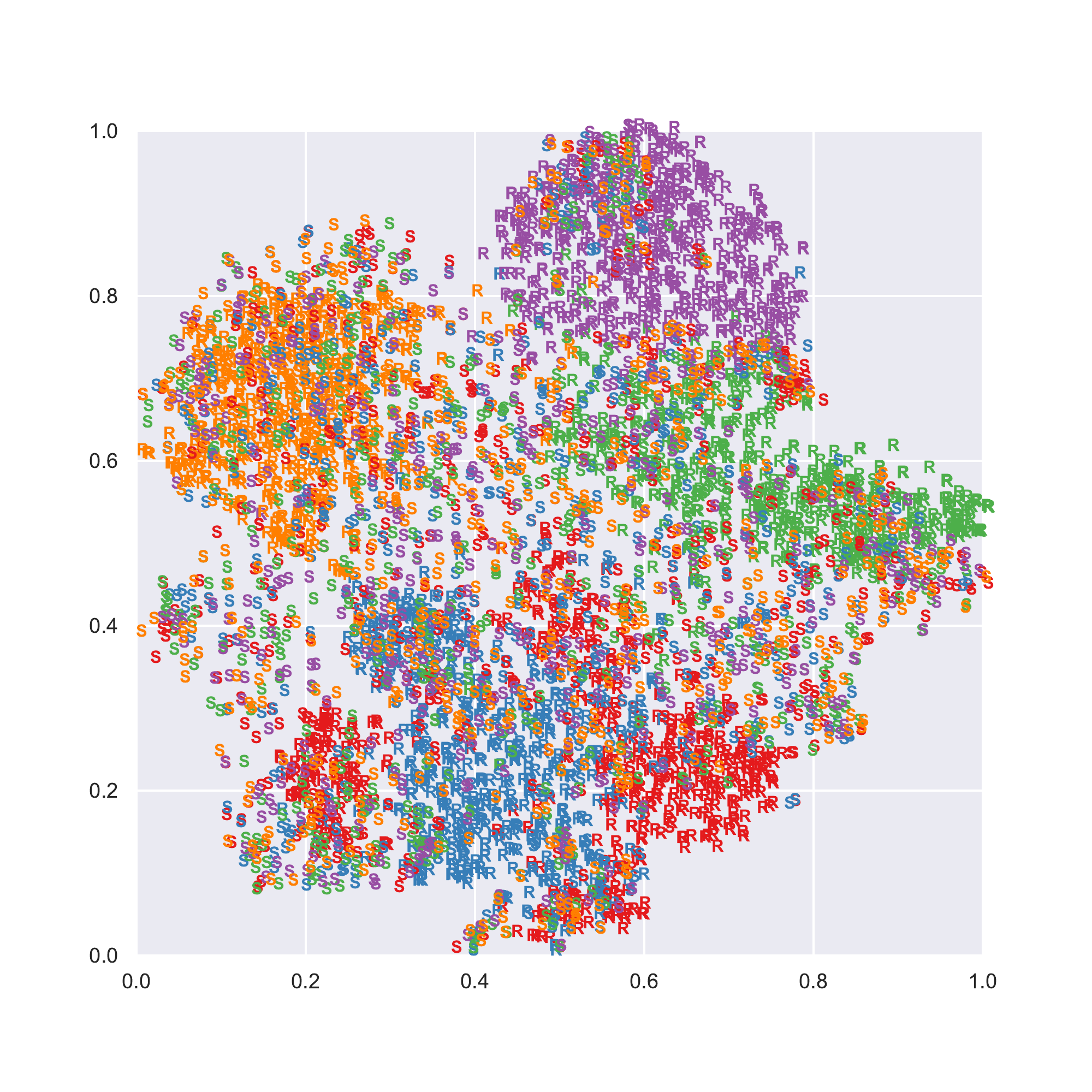}
    \caption{ABD}
    \label{fig:tsne_abd}
  \end{subfigure}
  \hfill
  \begin{subfigure}[b]{0.24\linewidth}
    \includegraphics[width=\linewidth]{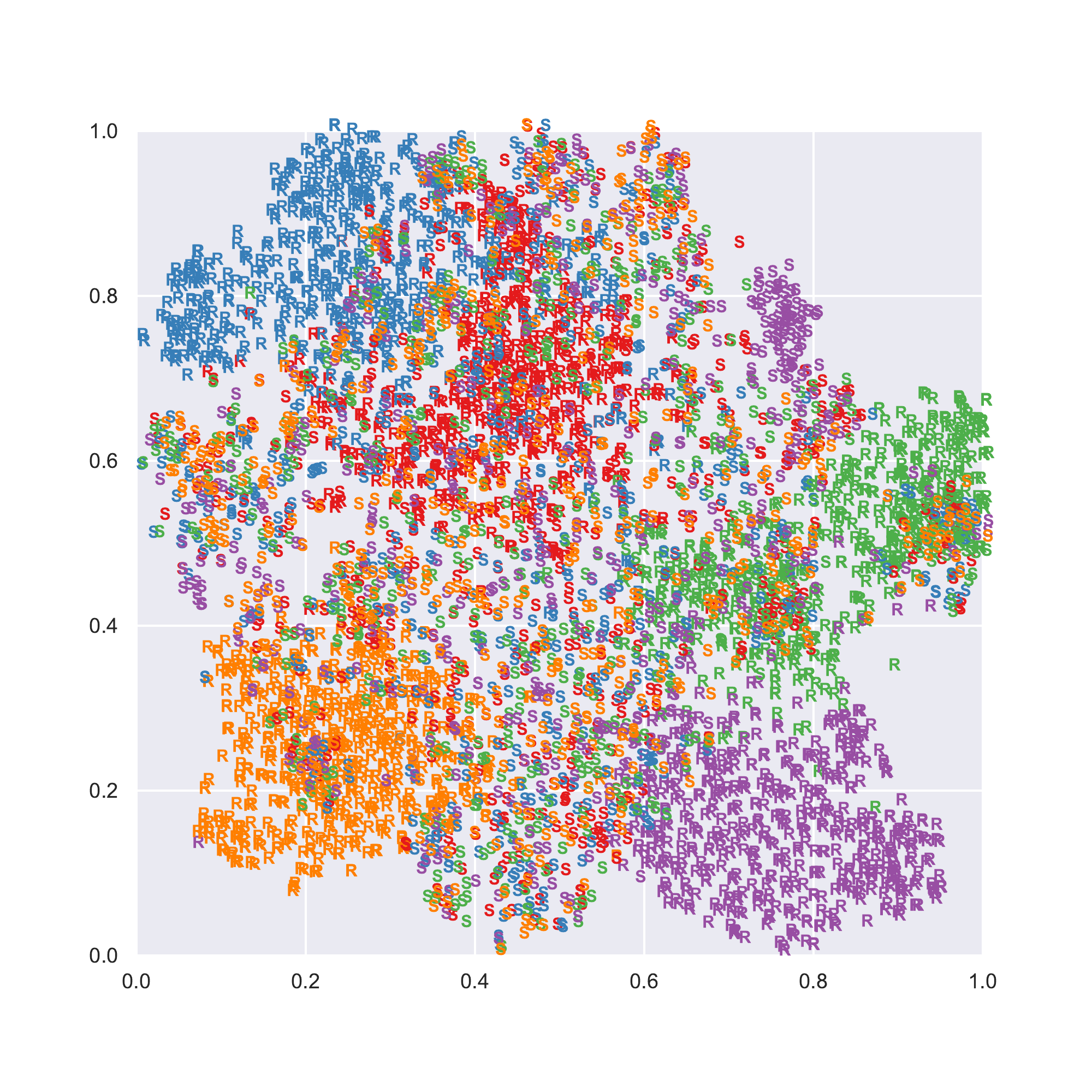}
    \caption{Ours}
    \label{fig:tsne_ours}
  \end{subfigure}
  \caption{\textbf{Visualization of t-SNE on synthetic data and real data}. ``R'' and ``S'' represent real and synthetic. Please zoom in. Same class have same color. The classes are randomly sampled. (a) jointly and (b)  independently reduce dimensions of synthetic data and real data of the first task, respectively, showing impacts of semantic  and domain  features, respectively. (c) and (d) are feature distributions of ABD's and ours final models, respectively.
  }
  \label{fig:tsne}
  \vspace{-20pt}
\end{figure}

\section{Additional Experimental Details}
\label{sec:exp}

\textbf{Data Augmentation.} 
We follow prior works~\cite{ucir,podnet,abd} to augment data in all of our experiments. For CIFAR100~\cite{cifar}, we first normalize images with means 0.5071, 0.4867, 0.4408 and standard variations 0.2675, 0.2565, 0.2761, then perform random crop with padding 4 and random horizontal flip with probability 0.5. For Tiny-ImageNet200~\cite{tiny_imagenet}, the normalization means and standard variations are  0.4803, 0.4481, 0.3976 and 0.2764, 0.2688, 0.2816, and the others are same with CIFAR100. As for ImageNet100~\cite{ucir}, we first resize images into $256 \times 256$, then crop the images into $224 \times 224$ from the center and normalize them with means 0.485, 0.456, 0.406 and standard variations 0.229, 0.224, 0.225. After that, we apply the same random crop and random horizontal flip augmentations as for the other two datasets.

\smallskip\noindent\textbf{Image Synthesis Implementation Details.}
As we presented in Sec. 3.4 of the main paper, there are four optimization objectives (\ie, label diversity, data content, stat alignment and image prior) for training the synthesizer, scale factors of which are set to 1,1,5 and 0.001 in all experiments.
The temperature parameter $\alpha_{temp}$ in label diversity loss is set to 1000. The number of training steps are 5000 for CIFAR100~\cite{cifar} and Tiny-ImageNet200~\cite{tiny_imagenet}, and 10000 for ImageNet100~\cite{ucir}.

\smallskip\noindent\textbf{Hyperparameter Tuning.}
We empirically set $\lambda_{lce}$ to 0.5 and tuned the other two parameters by a simple grid search from $\{0.05, 0.5, 5.0\}$ on CIFAR100 with $N=5$. Then, we found that the performance was insensitive to the value of $\lambda_{rkd}$, but the performance collapsed with the largest $\lambda_{hkd}=5.0$. So, we further searched $\lambda_{hkd}$ from $\{0.25, 0.15\}$, and the model performed better with 0.15. Finally, we conducted all other experiments on three datasets with $\lambda_{lce}=0.5$, $\lambda_{hkd}=0.15$ and $\lambda_{rkd}=0.5$, which verified their effectiveness and generality.

\section{Additional Experimental Results}
\label{sec:exp_res}

\smallskip\noindent\textbf{Additional Incremental Accuracy Plots.}
We depicted the 20-tasks and 26-tasks task-by-task incremental accuracy in Fig. 2 of the main paper. Here we provide more task-by-task incremental accuracy plots in Fig.~\ref{fig:cifar_inc_acc},~\ref{fig:tiny_inc_acc_sup} of this supplementary document. For the evaluation protocol introduced by Hou~\etal~\cite{ucir}, we implement the Data-Free UCIR (UCIR-DF)~\cite{ucir} and PODNet (PODNet-DF)~\cite{podnet}, and present their performances in Fig.~\ref{fig:cifar_inc_1_2},~\ref{fig:cifar_inc_2_2},~\ref{fig:cifar_inc_3_2}.

\smallskip\noindent\textbf{Ablation of Knowledge Distillation.}
In the Fig. 3 of the main paper, we ablate the RKD and HKD by directly removing them. We also conducted experiments that replace one with the other. The $\bar{A}_{20}$/$A_{20}$ drops from 49.47\%/30.92\% to 40.17\%/21.35\% for a single run when RKD is replaced with HKD, and decreases to 22.19\%/5.81\% when HKD is taken place by RKD. All these ablation studies demonstrate the importance of applying appropriate knowledge distillation methods to synthetic old and real new data.

\smallskip\noindent\textbf{Plasticity and Stability Comparison with ABD.}
In 5-tasks CIFAR100 experiments, we tuned the KD weights to  compare model's plasticity (based on the last accuracy) with controlled stability. Our R-DFCIL is higher than ABD by $7\%$ in last accuracy on the new task (better plasticity), with $4\%$ higher last accuracy on old tasks (also better stability). 

\smallskip\noindent\textbf{Newer Relational Knowledge Distillation.}
We also investigated some newer KD methods, such as CRD~\cite{crd} and CRCD~\cite{crcd}. These methods are more complex and do not solve the DFCIL problem as effectively as RKD. In our experiments, using RKD achieves an average accuracy gain of $5\%$ (CRCD) and $10\%$ (CRD).

\section{Feature Analysis and Bottlenecks in Prior Approaches}
\label{sec:fa}

In Sec. 1 of the main paper, we identified two bottlenecks in previous DFCIL approaches according to our study on feature analysis. Here we define semantic features as the discriminate features between classes and domain features as the features shared by a kind of data (real or synthetic). As shown in Fig.~\ref{fig:tsne_joint}, the synthetic and real data of the same class cluster together because the semantic features dominate domain features when jointly reducing dimensions. On the contrary, semantic features are dominated by domain features when independently reducing the dimensions of synthetic and real data (Fig.~\ref{fig:tsne_independent}), resulting in the mixing of the synthetic data that belong to different classes. Figure~\ref{fig:tsne_abd} and \ref{fig:tsne_ours} show that the overlap between real and synthetic data in ABD is more severe than ours, and the clustering of real data in ABD is worse than ours, indicating that our R-DFCIL learned more semantic features, \ie, we address the first bottleneck with local classification loss during representation learning. 

The figures also illustrate the different decision boundaries between classes in synthetic and real data. Our R-DFCIL is more robust to disturbed decision boundaries in synthetic data because we solve the second bottleneck by applying different knowledge distillation on real and synthetic data, effectively alleviating the conflict between the model's plasticity and stability.
\end{document}